\documentclass[journal]{IEEEtran}

\usepackage[numbers]{natbib}
\usepackage{booktabs}
\usepackage{caption}
\usepackage{verbatim}
\usepackage{lineno}
\usepackage{hyperref}
\usepackage{empheq}
\usepackage{url}
\usepackage{amsmath}
\usepackage{graphicx}
\usepackage[flushleft]{threeparttable}
\usepackage{subcaption}
\usepackage{booktabs}

\usepackage{amsmath,amssymb}
\graphicspath {{figure/}}

\begin{document}
\bstctlcite{IEEEexample:BSTcontrol}

\title{Evaluate the Malignancy of Pulmonary Nodules Using the 3D Deep Leaky Noisy-or Network}

    \author{Fangzhou~Liao,
        Ming Liang,
        Zhe Li,
        Xiaolin~Hu*,~\IEEEmembership{Senior Member,~IEEE}
        and~Sen~Song*
       
        \thanks{Fangzhou Liao, Zhe Li and Sen Song are with the School of Medicine, Tsinghua University, Beijing 100084, China.}
        
        \thanks{Ming Liang and Xiaolin Hu are with the State Key Laboratory of Intelligent Technology and Systems, Tsinghua
            National Laboratory for Information Science and Technology (TNList),
            and Department of Computer Science and Technology, Tsinghua
            University, Beijing 100084, China. (email: xlhu@tsinghua.edu.cn).
        }
        \thanks{This work was supported in part by the National Basic Research Program (973 Program) of China under grant no. 2013CB329403, the National Natural Science Foundation of China under grant nos. 91420201, 61332007, 61621136008 and 61620106010.}
         \thanks{* Corresponding authors}
    }

\markboth{Journal of \LaTeX\ Class Files,~Vol.~14, No.~8, August~2015}%
{Shell \MakeLowercase{\textit{et al.}}: Bare Demo of IEEEtran.cls for IEEE Journals}

\maketitle

\begin{abstract}
Automatic diagnosing lung cancer from Computed Tomography (CT) scans involves two steps: detect all suspicious lesions (pulmonary nodules) and evaluate the whole-lung/pulmonary malignancy. Currently, there are many studies about the first step, but few about the second step. 
Since the existence of nodule does not definitely indicate cancer, and the morphology of nodule has a complicated relationship with cancer, the diagnosis of lung cancer demands careful investigations on every suspicious nodule and integration of information of all nodules. 
We propose a 3D deep neural network to solve this problem. 
The model consists of two modules. The first one is a 3D region proposal network for nodule detection, which outputs all suspicious nodules for a subject. The second one selects the top five nodules based on the detection confidence, evaluates their cancer probabilities and combines them with a leaky noisy-or gate to obtain the probability of lung cancer for the subject. The two modules share the same backbone network, a modified U-net. The over-fitting caused by the shortage of training data is alleviated by training the two modules alternately.
The proposed model won the first place in the Data Science Bowl 2017 competition. The code has been made publicly available\footnote{https://github.com/lfz/DSB2017}.
\end{abstract}


%
\IEEEpeerreviewmaketitle

\section{Introduction}

\IEEEPARstart{L}{ung} cancer is one of the most common and deadly malignant cancers. Like other cancers, the best solution for lung cancer is early diagnosis and timely treatment. Therefore regular examinations are necessary.
The volumetric thoracic Computed Tomography (CT) is a common imaging tool for lung cancer diagnosis \cite{infante_randomized_2009}. It visualizes all tissues according to their absorption of X-ray. The lesion in the lung is called pulmonary nodules. A nodule usually has the same absorption level as the normal tissues, but has a distinctive shape: the bronchus and vessels are continuous pipe systems, thick at the root and thin at the branch, and nodules are usually spherical and isolated. It usually takes an experienced doctor around 10 minutes to perform a thorough check for a patient, because some nodules are small and hard to be found. Moreover, there are many subtypes of nodules, and the cancer probabilities of different subtypes are different. Doctors can evaluate the malignancy of nodules based on their morphology, but the accuracy highly depends on doctors' experience, and different doctors may give different predictions \cite{singh2012reader}. 

Computer-aided diagnosis (CAD) is suitable for this task because computer vision models can quickly scan everywhere with equal quality and they are not affected by fatigue and emotions. Recent advancement of deep learning has enabled computer vision models to help the doctors to diagnose various problems and in some cases the models have exhibited competitive performance to doctors \cite{shin2013stacked,esteva2017dermatologist,gulshan_development_2016,litjens_survey_2017,duncan2000medical}. 

\begin{figure*}[!htbp]
    \centering
    \includegraphics[width=\linewidth]{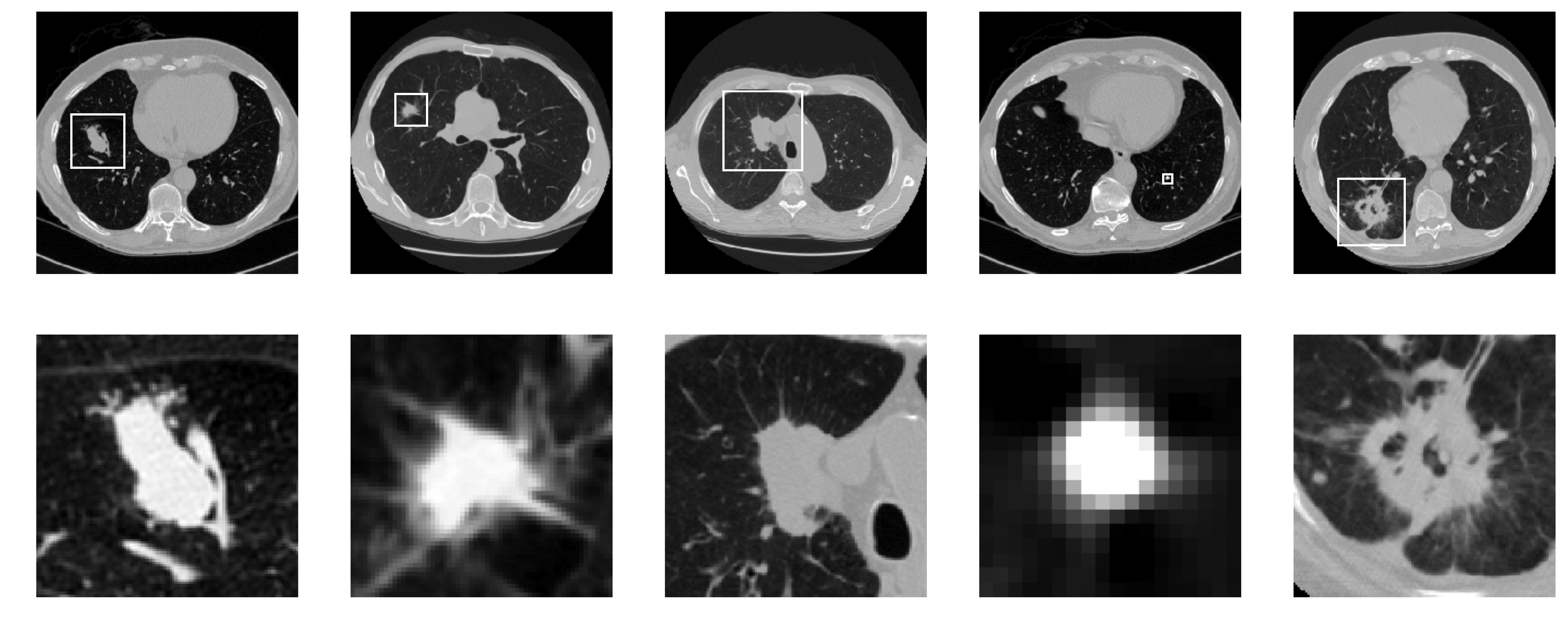}
    \caption{Examples of nodules in the DSB dataset. Top: the whole slice. Bottom: the zoomed image.}
    \label{fig_examples}
\end{figure*}

Automatic lung cancer diagnosing has several difficulties compared with general computer vision problems. First, nodule detection is a 3D object detection problem which is harder than 2D object detection. Direct generalization of 2D object detection methods to 3D cases faces technical difficulty due to the limited GPU memory. Therefore some methods use 2D region proposal networks (RPN) to extract proposals in individual 2D images then combine them to generate 3D proposals \cite{peng2016multi,ding_accurate_2017}. More importantly, labeling 3D data is usually much harder than labeling 2D data, which may make deep learning models fail due to over-fitting. Second, the shape of the nodules is diverse (Fig. \ref{fig_examples}), and the difference between nodules and normal tissues is vague. In consequence, even experienced doctors cannot reach consensuses in some cases \cite{sg_3rd_assessment_2009}. Third, the relationship between nodule and cancer is complicated. The existence of nodule does not definitely indicate lung cancer. For patients with multiple nodules, all nodules should be considered to infer the cancer probability. In other words, unlike the classical detection task and the classical classification task, in this task, a label corresponds to several objects. This is a multiple instance learning (MIL) \cite{dietterich_solving_1997} problem, which is a hard problem in computer vision.

To tackle these difficulties, we take the following strategies. We built a 3D RPN \cite{ren2015faster} to directly predict the bounding boxes for nodules. The 3D convolutional neural network (CNN) structure enables the network to capture complex features. To deal with the GPU memory problem, a patch-based training and testing strategy is used. The model is trained end-to-end to achieve efficient optimization. Extensive data augmentation is used to combat over-fitting. The threshold for the detector is set low such that all suspicious nodules are included. Then the top five suspicious nodules are selected as input to the classifier. A leaky noisy-or model \cite{pearl2014probabilistic} is introduced in the classifier to combine the scores of top five nodules.

The noisy-or model is a local causal probability model commonly used in probability graph models \cite{pearl2014probabilistic}. It assumes that an event can be caused by different factors, and the happening of any one of these factors can lead to the happening of the event with independent probability. One modified version of the model is called leaky noisy-or model \cite{pearl2014probabilistic}, which assumes that there is a leakage probability for the event even none of the factors happens.
The leaky noisy-or model is suitable for this task. First, when multiple nodules are present in a case, all nodules contribute to the final prediction. Second, a highly suspicious nodule would explain away the cancer case, which is desirable. Third, when no nodule can explain a cancer case, cancer can be attributed to a leakage probability. 

The classification network is also a 3D neural network. To prevent over-fitting, we let the classification network share the backbone of the detection network (the parameters of the backbones of the two networks are tied) and train the two networks alternately. Extensive data augmentation are also used. 

Our contributions in this work are summarized as follows: 
\begin{enumerate}
\item To the best of our knowledge, we propose the first volumetric one-stage end-to-end CNN for 3D object detection. 
\item We propose to integrate the noisy-or gate into neural networks to solve the multi-instance learning task in CAD.
\end{enumerate}
We validated the proposed method on the Data Science Bowl 2017\footnote{https://www.kaggle.com/c/data-science-bowl-2017} and won the first place among 1972 teams.

The rest of the paper is organized as follows. Section \ref{sec:related} presents some closely related works. The pipeline of the proposed method is detailed in subsequent sections. It consists of three steps: (1) preprocessing (Section \ref{sec:datasets}): segment the lung out from other tissues; (2) detection (Section \ref{sec:detection}): find all suspicious nodules in the lung; (3) classification (Section \ref{sec:classification}): score all nodules and combine their cancer probabilities to get the overall cancer probability of the patient. The first step is accomplished by classical image preprocessing techniques and the other two steps by neural networks. The results are presented in Sections \ref{sec:results}. Section \ref{sec:discussion} concludes the paper with some discussions.

\section{Related works}\label{sec:related}
\subsection{General object detection}
A number of object detection methods have been proposed and a thorough review is beyond the scope of this paper. Most of these methods are designed for 2D object detection.
Some state-of-the-art methods have two stages (e.g., Faster-RCNN \cite{ren2015faster}), in which some bounding boxes (called {\it proposals}) are proposed in the first stage (containing an object or not) and the class decision (which class the object in a proposal belongs to) is made in the second stage. More recent methods have a single stage, in which the bounding boxes and class probabilities are predicted simultaneously (YOLO \cite{redmon2016yolo9000}) or the class probabilities are predicted for default boxes without proposal generation (SSD \cite{liu2016ssd}). In general, single-stage methods are faster but two-stage methods are more accurate. In the case of single class object detection, the second stage in the two-stage methods is no longer needed and the methods degenerate to single-stage methods. 


Extension of the cutting-edge 2D object detection methods to 3D object detections tasks (e.g., action detection in video and volumetric detection) is limited. 
Due to the memory constraint in mainstream GPUs, some studies use 2D RPN to extract proposals in individual 2D images then use an extra module to combine the 2D proposal into 3D proposals \cite{peng2016multi,ding_accurate_2017}. Similar strategies have been used for 3D image segmentation \cite{chen2016combining}. As far as we know 3D RPN has not been used to process video or volumetric data.

\subsection{Nodule detection}
Nodule detection is a typical volumetric detection task. Due to its great clinical significance, it draws more and more attention in these years. This task is usually divided into two subtasks \cite{van_ginneken_comparing_2010}: making proposals and reducing false positives, and each subtask has attracted many researches. The models for the first subtask usually start with a simple and fast 3D descriptor then followed by a classifier to give many proposals. The models for the second subtask are usually complex classifiers. In 2010 \citet{van_ginneken_comparing_2010} gave a comprehensive review of six conventional algorithms and evaluated them on the ANODE09 dataset, which contains 55 scans. During 2011-2015, a much larger dataset LIDC \cite{armato_iii_samuel_g._data_2015,armato_lung_2011,clark_cancer_2013} was developed. Researchers started to adopt CNN to reduce the number of false positives. \citet{setio_pulmonary_2016} adopted a multi-view CNN, and \citet{dou_multi-level_2016} adopted a 3D CNN to solve this problem and both achieved better results than conventional methods. \citet{ding_accurate_2017} adopted 2D RPN to make nodule proposals in every slice and adopted 3D CNN to reduce the number of false-positive samples. A competition called LUng Nodule Analysis 2016 (LUNA16) \cite{setio_validation_2016} was held based on a selected subset of LIDC. In the detection track of this competition, most participants used the two-stage methods \cite{setio_validation_2016}.

\subsection{Multiple instance learning}
In MIL task, the input is a bag of instances. The bag is labeled positive if any of the instances are labeled positive and the bag is labeled negative if all of the instances are labeled negative. 


Many medical image analysis tasks are MIL tasks, so before the rise of deep learning, some earlier works have already proposed MIL frameworks in CAD. \citet{dundar_multiple-instance_2008} introduced convex hull to represent multi-instance features and applied it to pulmonary embolism and colon cancer detection. 
\citet{xu2014deep} extracted many patches from the tissue-examing image and treated them as multi-instances to solve the colon cancer classification problem.

To incorporate the MIL into deep neural network framework, the key component is a layer that combines the information from different instances together, which is called MIL Pooling Layer (MPL \cite{wang2016revisiting}).  Some MPL examples are: max-pooling layer \cite{wu2015deep}, mean pooling layer \cite{wang2016revisiting}, log-sum-exp pooling layer \cite{pinheiro2015image}, generalized-mean layer \cite{xu2014deep} and noisy-or layer \cite{sun_multiple_2016}. If the number of instances is fixed for every sample, it is also feasible to use feature concatenation as an MPL \cite{zeng2015deep}. The MPL can be used to combine different instances in the feature level \cite{wu2015deep,pinheiro2015image} or output level \cite{sun_multiple_2016}.

\subsection{Noisy-or model}
The noisy-or Bayesian model is wildly used in inferring the probability of diseases such as liver disorder \cite{onisko2001learning} and asthma case \cite{anand2008probabilistic}. \citet{heckerman1990tractable} built a multi-features and multi-disease diagnosing system based on the noisy-or gate. \citet{halpern2013unsupervised} proposed an unsupervised learning method based on the noisy-or model and validated it on the Quick Medical Reference model. 

All of the studies mentioned above incorporate the noisy-or model into the Bayesian models. Yet the integration of the noisy-or model and neural networks is rare. \citet{sun_multiple_2016} has adopted it as an MPL in the deep neural network framework to improve the image classification accuracy. And \citet{zhang2006multiple} used it as a boosting method to improve the object detection accuracy. 

\section{Datasets and preprocessing}\label{sec:datasets}
\subsection{Datasets}
Two lung scans datasets are used to train the model, the LUng Nodule Analysis 2016 dataset (abbreviated as LUNA) and the training set of Data Science Bowl 2017 (abbreviated as DSB). The LUNA dataset includes 1186 nodule labels in 888 patients annotated by radiologists, while the DSB dataset only includes the per-subject binary labels indicating whether this subject was diagnosed with lung cancer in the year after the scanning. The DSB dataset includes 1397, 198, 506 persons (cases) in its training, validation, and test set respectively. We manually labeled 754 nodules in the training set and 78 nodules in the validation set.

\begin{figure}
\centering
    \begin{subfigure}[t]{0.75\columnwidth}
    \includegraphics[width=\columnwidth]{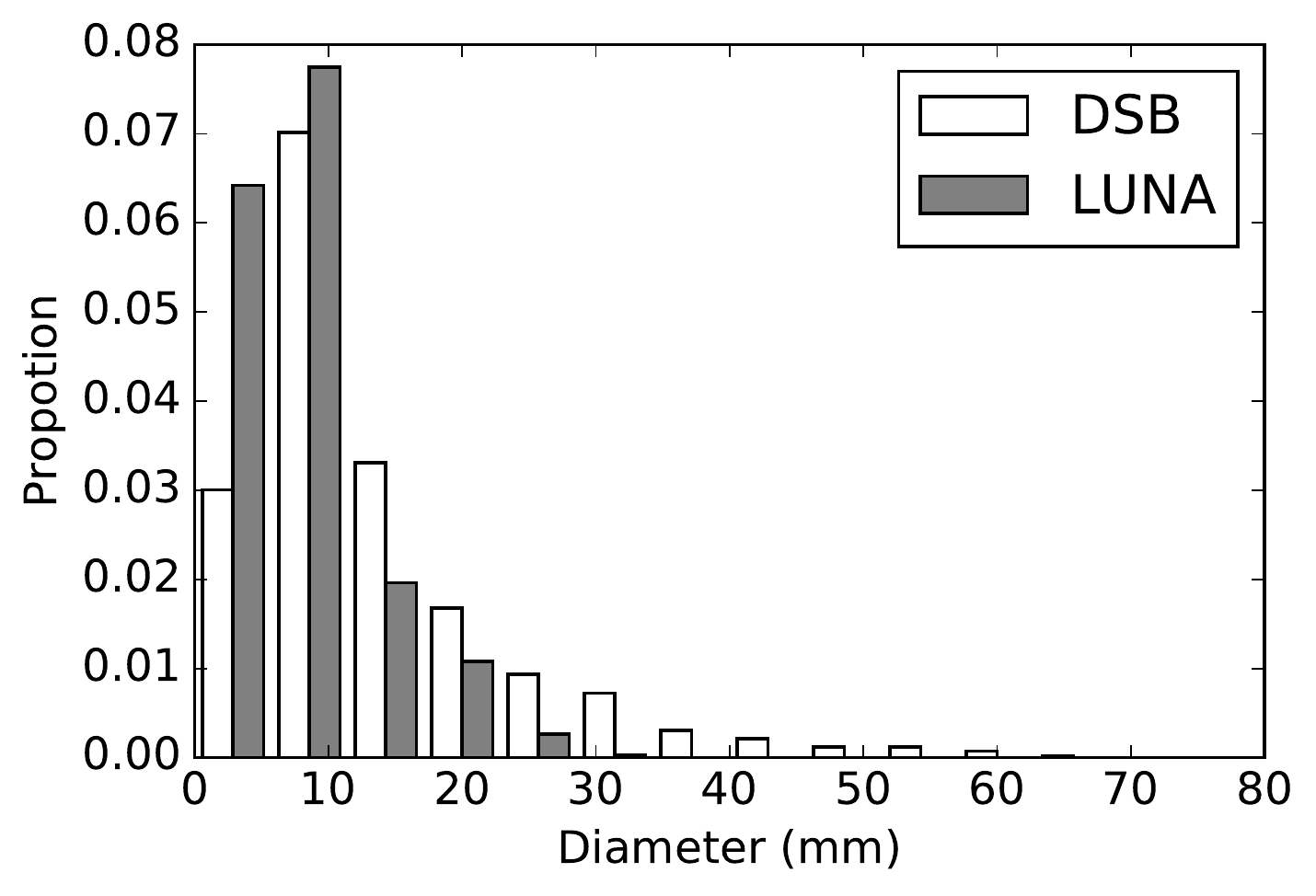}
    \subcaption{}
    \label{fig_rdist1}
    \end{subfigure}
    \begin{subfigure}[t]{0.75\columnwidth}
    \includegraphics[width=\columnwidth]{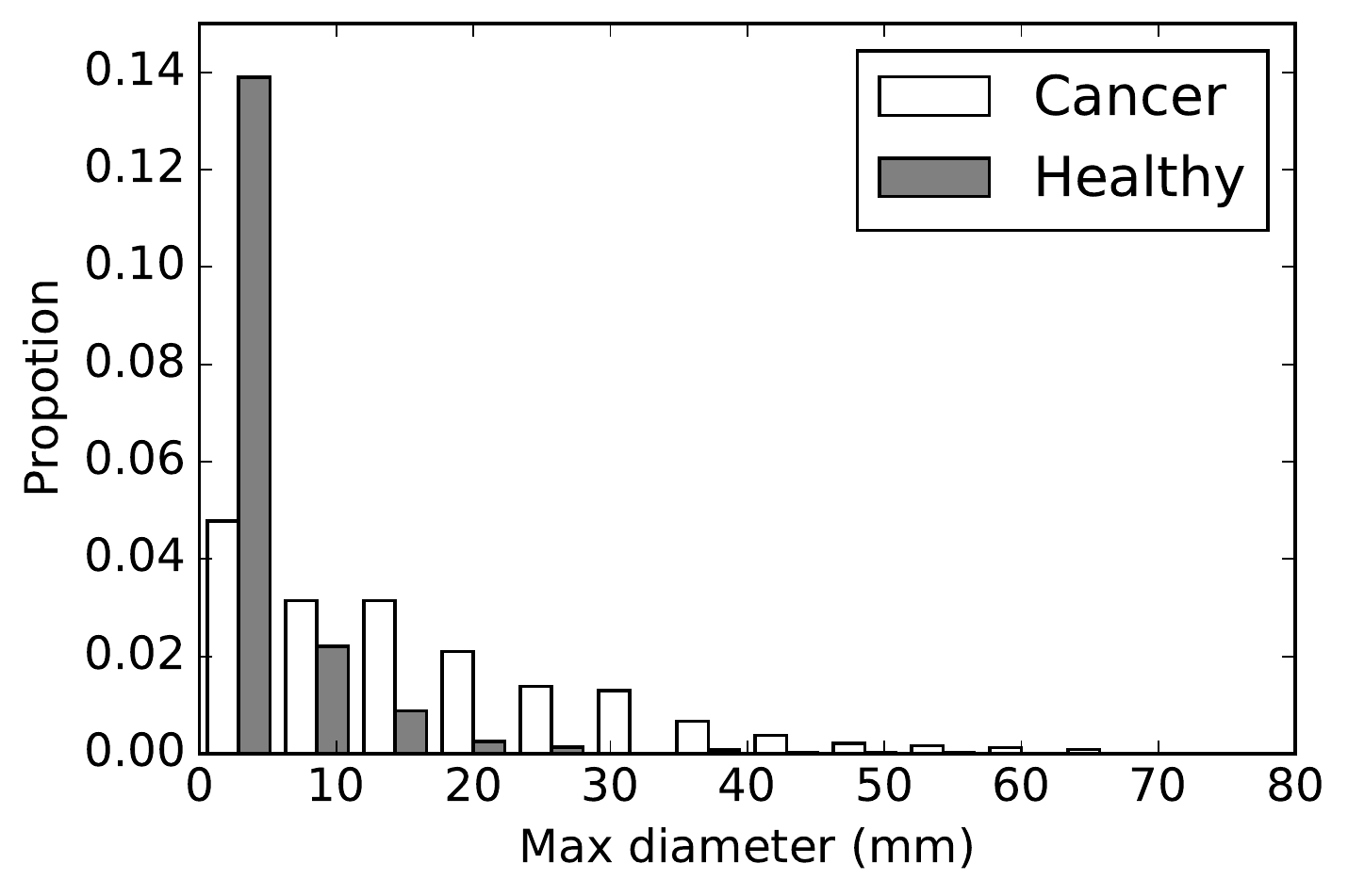}
    \subcaption{}
    \label{fig_rdist2}
    \end{subfigure}
    \caption{Distributions of the nodule diameter. (a) Distributions in the DSB and LUNA datasets. (b) Distributions of the maximum nodule diameter for cancer patient and healthy people in the DSB dataset.}
\end{figure}

There are some significant differences between LUNA nodules and DSB nodules. The LUNA dataset has many very small annotated nodules, which may be irrelevant to cancer. According to doctors' experience \cite{macmahon2017guidelines}, the nodules smaller than 6 mm are usually not dangerous. However, the DSB dataset has many very big nodules (larger than 40 mm) (the fifth sample in Fig. 1). The average nodule diameter is 13.68 mm in the DSB dataset and 8.31 mm in the LUNA dataset (Fig. \ref{fig_rdist1}). In addition, the DSB dataset has many nodules on the main bronchus (third sample in Fig. 1), which are rarely found in the LUNA dataset. If the network is trained on the LUNA dataset only, it will be difficult to detect the nodules in the DSB dataset. Missing big nodules would lead to incorrect cancer predictions as the existence of big nodules is a hallmark of cancer patients (Fig. \ref{fig_rdist2}). To cope with these problems, we remove the nodules smaller than 6 mm from LUNA annotations and manually labeled the nodules in DSB.

The authors have no professional knowledge of lung cancer diagnosis, so the nodule selection and manual annotations may raise considerable noise. The model in the next stage (cancer classification) is designed to be robust to wrong detections, which alleviates the demand for highly reliable nodule labels. 

\subsection{Preprocessing}
The overall preprocessing procedure is illustrated in Fig. \ref{fig_prep}.
All raw data are firstly converted into Hounsfield Unit (HU), which is a standard quantitative scale for describing radiodensity. Every tissue has its own specific HU range, and this range is the same for different people (Fig. \ref{fig_prep}a).

\subsubsection{Mask extraction}
A CT image contains not only the lung but also other tissues, and some of them may have spherical shapes and look like nodules. To rule out those distractors, the most convenient method is extracting the mask of lung and ignore all other tissues in the detection stage.
For each slice, the 2D image is filtered with a Gaussian filter (standard deviation = 1 pixel) and then binarized using -600 as the threshold (Fig. \ref{fig_prep}b). All 2D connected components smaller than 30 mm\textsuperscript{2} or having eccentricity greater than 0.99 (which correspond to some high-luminance radial imaging noise) are removed. Then all 3D connected components in the resulting binary 3D matrix are calculated, and only those not touching the matrix corner and having a volume between 0.68 L and 7.5 L are kept. 

After this step, usually there is only one binary component left corresponding to the lung, but sometimes there are also some distracting components. Compared with those distracting components, the lung component is always at the center position of the image. For each slice of a component, we calculate the minimum distance from it to the image center (MinDist) and its area. Then we select all slices whose area $>6000~\text{mm}^2$ in the component, and calculate the average MinDist of these slices. If the average MinDist is greater than $62~\text{mm}$, this component is removed. The remaining components are then unioned, representing the lung mask (Fig. \ref{fig_prep}c).

The lung in some cases is connected to the outer world on the top slices, which makes the procedure described above fail to separate the lung from the outer world space. Therefore these slices need to be removed first to make the above processing work.

\subsubsection{Convex hull \& dilation}
There are some nodules attached to the outer wall of the lung. They are not included in the mask obtained in the previous step, which is unwanted. To keep them inside the mask, a convenient way is to compute the convex hull of the mask. Yet directly computing the convex hull of the mask would include too many unrelated tissues (like the heart and spine). So the lung mask is first separated into two parts (approximately corresponding to the left and right lungs) before the convex hull computation using the following approach. 

\begin{figure}
\centering
\includegraphics[width = \linewidth]{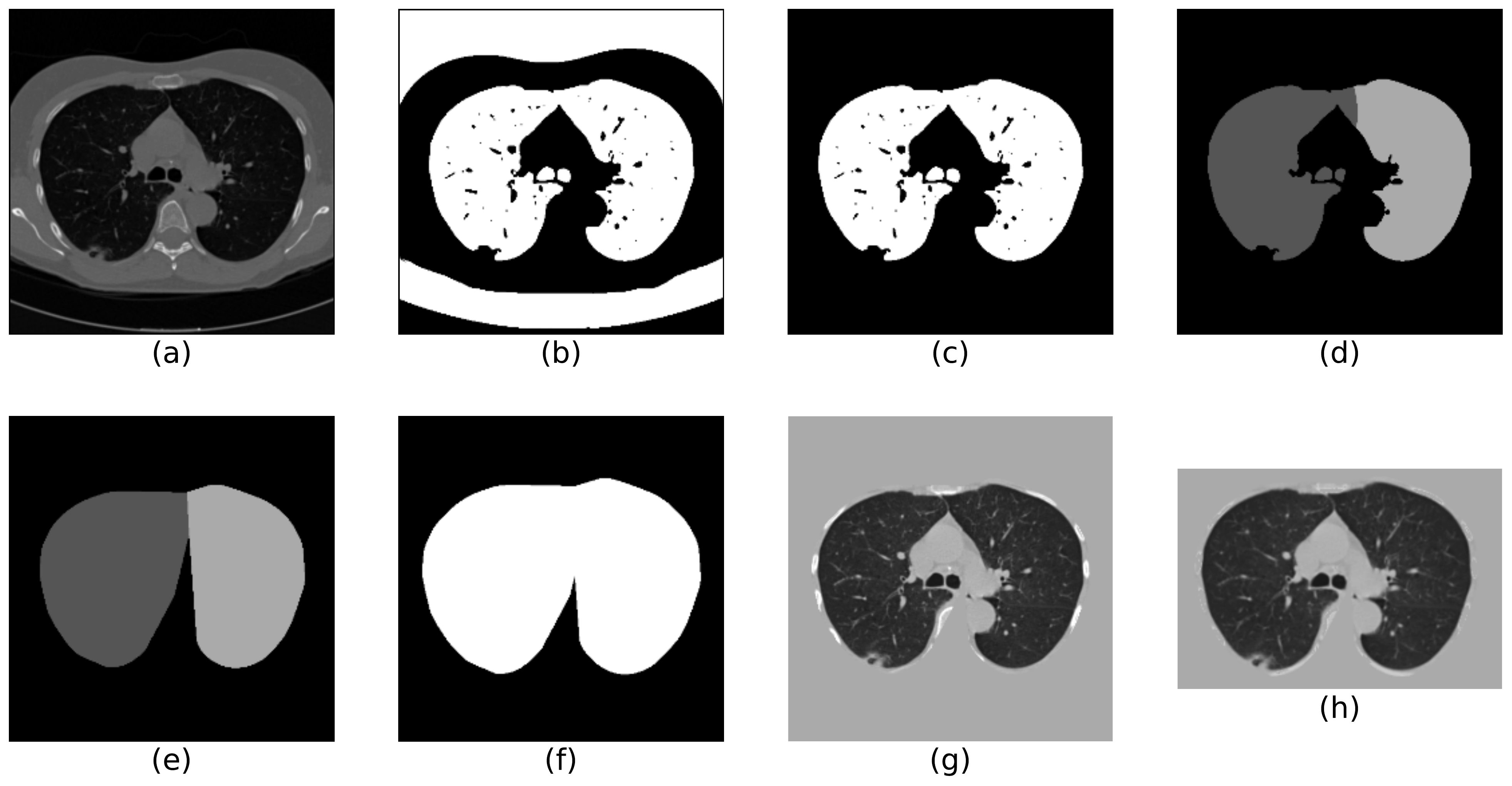}
\caption{The procedures of preprocessing. Notice the nodule sticking to the outer wall of lungs. (a) Convert the image to HU, (b) binarize image by thresholding, (c) select the connected domain corresponding to the lungs, (d) segment the left and right lungs, (e) compute the convex hull of each lung. (f) dilate and combine the two masks, (g) multiply the image with the mask, fill the masked region with tissue luminance, and convert the image to UINT8, (h) crop the image and clip the luminance of bone.}
\label{fig_prep}

\centering
\includegraphics[width = \linewidth]{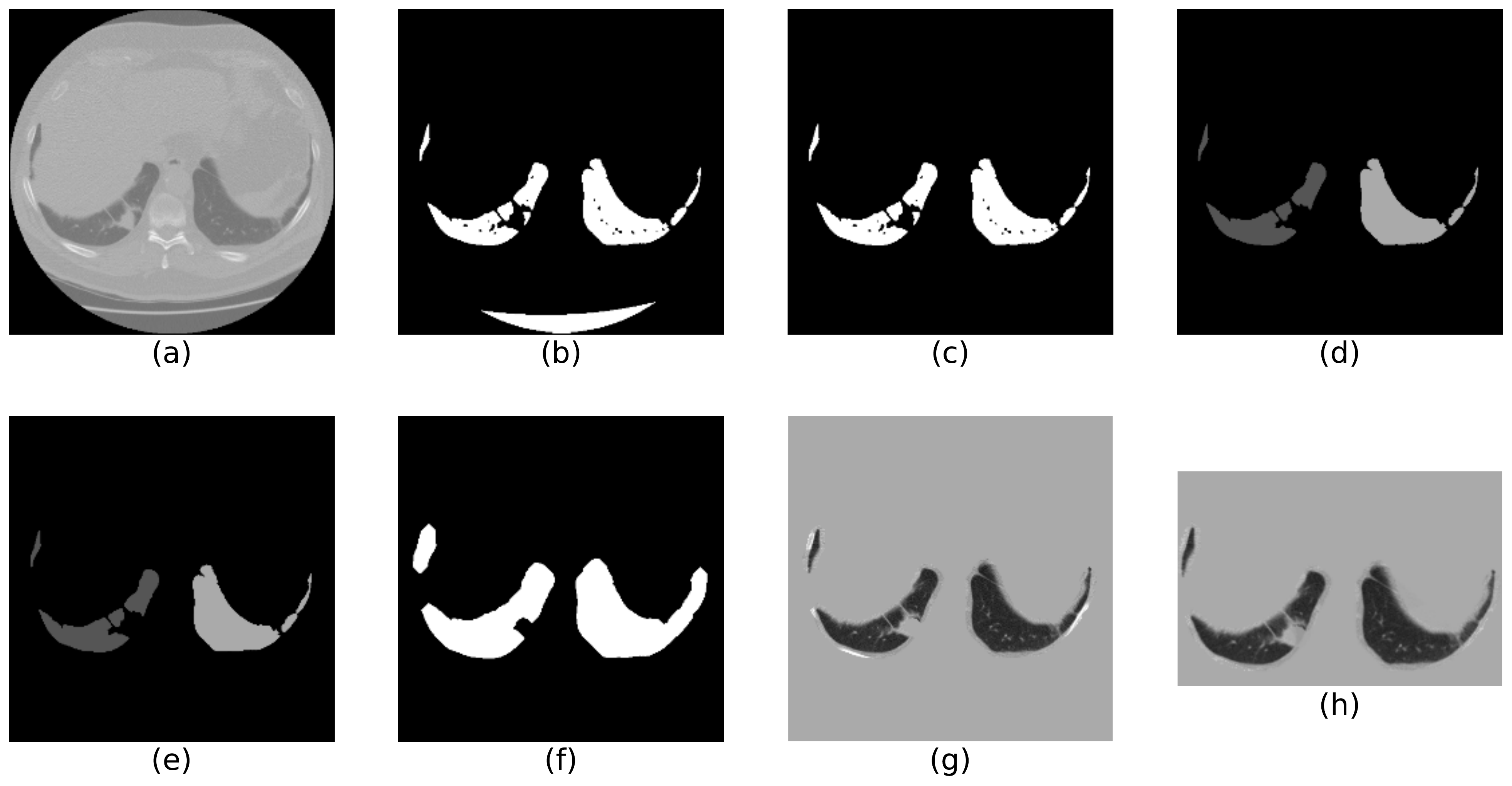}
\caption{The same as Fig. \ref{fig_prep}, but a lower slice is shown. Notice that no convex hull is calculated in step (e).}
\label{fig_prep2}

\end{figure}

The mask is eroded iteratively until it is broken into two components (their volumes would be similar), which are the central parts of the left and right lungs. Then the two components are dilated back to original sizes. Their intersections with the raw mask are now masks for the two lungs separately (Fig. \ref{fig_prep}d). For each mask, most 2D slices are replaced with their convex hulls to include those nodules mentioned above (Fig. \ref{fig_prep}e). The resultant masks are further dilated by 10 voxels to include some surrounding space. A full mask is obtained by unioning the masks for the two lungs (Fig. \ref{fig_prep}f).  

However, some 2D slices of the lower part of the lung have crescent shapes (Fig. \ref{fig_prep2}). Their convex hulls may contain too many unwanted tissues. So if the area of the convex hull of a 2D mask is larger than 1.5 times that of the mask itself, the original mask is kept (Fig. \ref{fig_prep2}e). 


\subsubsection{Intensity normalization}
To prepare the data for deep networks, we transform the image from HU to UINT8. The raw data matrix is first clipped within [-1200, 600], and linearly transformed to [0, 255]. It is then multiplied by the full mask obtained above, and everything outside the mask is filled with 170, which is the luminance of common tissues. 
In addition, for the space generated by dilation in the previous step, all values greater than 210 are replaced with 170 too. Because the surrounding area contains some bones (the high-luminance tissues), they are easily misclassified as calcified nodules (also high-luminance tissues). We choose to fill the bones with 170 so that they look like normal tissue (Fig. \ref{fig_prep}g). The image is cropped in all 3 dimensions so that the margin to every side is 10 pixels (Fig. \ref{fig_prep}h).

\section{3D CNN for Nodule Detection}\label{sec:detection}

A 3D CNN is designed to detect suspicious nodules. It is a 3D version of the RPN using a modified U-net \cite{ronneberger_u-net_2015} as the backbone model. Since there are only two classes (nodule and non-nodule) in this task, the predicted proposals are directly used as detection results without an additional classifier. This is similar to the one-stage detection systems YOLO \cite{redmon2016yolo9000} and SSD \cite{liu2016ssd}. This nodule detecting model is called N-Net for short, where N stands for nodule.

\subsection{Patch-based input for training}
Object detection models usually adopt the image-based training approach: during training, the entire image is used as input to the models. However, this is infeasible for our 3D CNN due to the GPU memory constraint. When the resolution of lung scans is kept at a fine level, even a single sample consumes more than the maximum memory of mainstream GPUs.

To overcome this problem, small 3D patches are extracted from the lung scans and input to the network individually. The size of the patch is $128 \times 128 \times 128 \times 1$ ($Height \times Length \times Width \times Channel$, the same notation is used in what follows). Two kinds of patches are randomly selected. First, 70\% of the inputs are selected so that they contain at least one nodule. Second, 30\% of the inputs are cropped randomly from lung scans and may not contain any nodules. The latter kind of inputs ensures the coverage of enough negative samples.

If a patch goes beyond the range of lung scans, it is padded with value 170, same as in the preprocessing step. The nodule targets are not necessarily located at the center of the patch but had a margin larger than 12 pixels from the boundary of the patch (except for a few nodules that are too large).

Data augmentation is used to alleviate the over-fitting problem. The patches are randomly left-right flipped and resized with a ratio between 0.8 and 1.15. Other augmentations such as axes swapping and rotation are also tried but no significant improvement is yielded.

\subsection{Network structure}
\begin{figure*}
    \centering
    \begin{subfigure}[t]{0.8\linewidth}
    \includegraphics[width=\columnwidth]{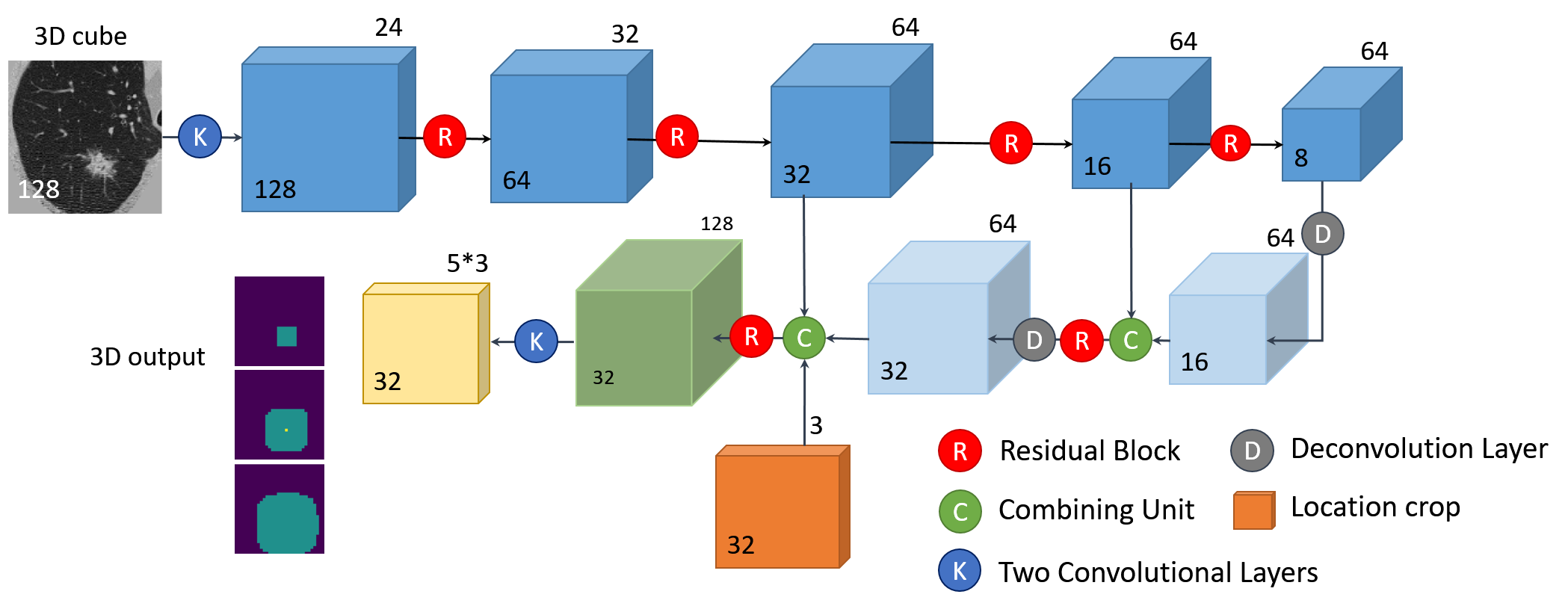}
    \subcaption{}
    \end{subfigure}
    
    \begin{subfigure}[t]{0.7\columnwidth}
    \includegraphics[width=\columnwidth]{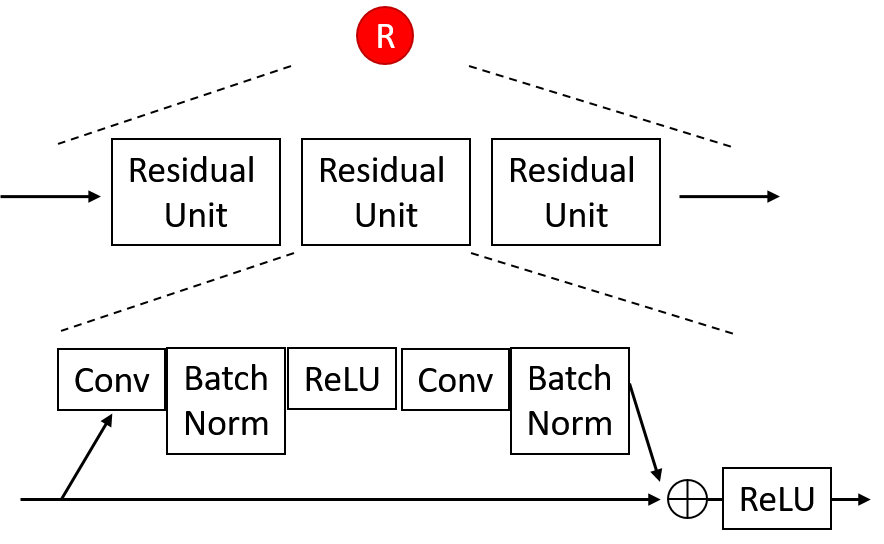}
    \subcaption{}
    \label{fig_resblock}
    \end{subfigure}
    \begin{subfigure}[t]{0.5\columnwidth}
    \includegraphics[width=\columnwidth]{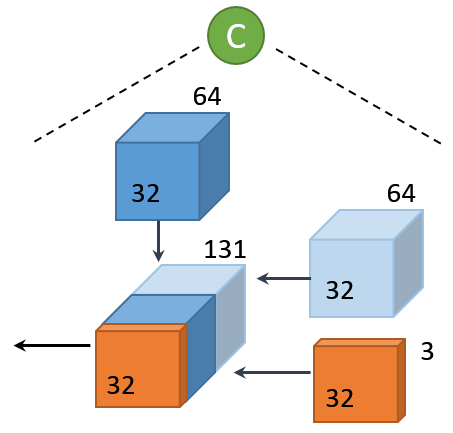}
    \subcaption{}
    \end{subfigure}

\caption{The nodule detection net. (a) The overall network structure. Each cube in the figure stands for a 4D tensor. Only two  dimensions are indicated in the figure. The number inside the cube stands for the spatial size ($Height=Width=Length$). The number outside the cube stands for the number of channels. (b) The structure of a residual block. (c) The structure of the left combining unit in (a). The structure of the right combining unit is similar but without the location crop.}
\label{Fig_N-Net}
\end{figure*}

The detector network consists of a U-Net \cite{ronneberger_u-net_2015} backbone and an RPN output layer, and its structure is shown in Fig. \ref{Fig_N-Net}. The U-Net backbone enabled the network to capture multi-scale information, which is essential because the size of nodules has large variations. The output format of RPN enables the network to generate proposals directly. 

The network backbone has a feedforward path and a feedback path (Fig. \ref{Fig_N-Net}a). The feedforward path starts with two $3\times 3 \times 3$ convolutional layers both with 24 channels. Then it is followed by four 3D residual blocks \cite{he_deep_2015} interleaved with four 3D max pooling layers (pooling size is $2\times 2\times 2$ and stride is 2). Each 3D residual block (Fig. \ref{fig_resblock}) is composed of three residual units \cite{he_deep_2015}. The architecture of the residual unit is illustrated in Fig. \ref{Fig_N-Net}b. All the convolutional kernels in the feedforward path have a kernel size of $3\times3\times3$ and a padding of 1.

The feedback path is composed of two deconvolutional layers and two combining units. Each deconvolutional layer has a stride of 2 and a kernel size of 2. And each combining unit concatenates a feedforward blob and a feedback blob and send the output to a residual block (Fig. \ref{Fig_N-Net}c). In the left combining unit, we introduce the location information as an extra input (see Section \ref{subsec:location} for details). The feature map of this combining unit has size $32\times32\times32\times131$. It is followed by two $1\times1\times1$ convolutions with channels 64 and 15 respectively, which results in the output of size $32\times32\times32\times15$. 

The 4D output tensor is resized to $32\times32\times32\times3\times5$. The last two dimensions correspond to the anchors and regressors respectively. Inspired by RPN, at every location, the network has three anchors of different scales, corresponding to three bounding boxes with the length of 10, 30, and 60 mm, respectively. So there are $32\times32\times32\times3$ anchor boxes in total. The five regression values are $(\hat{o},\hat{d}_x, \hat{d}_y,\hat{d}_z,\hat{d}_r)$. 
A sigmoid activation function is used for the first one:
\begin{equation*}
\hat{p} = \frac{1}{1+\exp({-\hat{o}})},
\end{equation*}
and no activation function is used for the others.

\subsection{Location information}\label{subsec:location}
The location of the proposal might also influence the judgment of whether it is a nodule and whether it is malignant, so we also introduce the location information in the network. For each image patch, we calculate its corresponding location crop, which is as big as the output feature map ($32\times32\times32\times3$). The location crop has 3 feature maps, which correspond to the normalized coordinates in X, Y, Z axis. In each axis, the maximal and minimal values in each axis are normalized to 1 and -1 respectively, which correspond to the two ends of the segmented lung.

\subsection{Loss function}
Denote the ground truth bounding box of a target nodule by $(G_x ,G_y, G_z, G_r)$ and the bounding box of an anchor by $(A_x, A_y, A_z, A_r)$, where the first three elements denote the coordinates of the center point of the box and the last element denotes the side length.
Intersection over Union (IoU) is used to determine the label of each anchor box. Anchor boxes whose IoU with the target nodule larger than 0.5 and smaller than 0.02 are treated as positive and negative samples, respectively. Others are neglected in the training process. The predicted probability and label for an anchor box is denoted by $\hat{p}$ and $p$ respectively. Note that $p\in\{0,1\}$ ($0$ for negative samples and $1$ for positive samples). The classification loss for this box is then defined by:
\begin{equation}
L_{cls} = p\log(\hat{p})+(1-p)\log(1-\hat{p}).
\end{equation}

The bounding box regression labels are defined as
\begin{align*}
d_x &= (G_x-A_x)/A_r,\\
d_y &= (G_y-A_y)/A_r,\\
d_z &= (G_z-A_z)/A_r,\\
d_r &= \log(G_r/A_r).
\end{align*}
The corresponding predictions are $\hat{d}_x,\hat{d}_y,\hat{d}_z,\hat{d}_r$, respectively.
The total regression loss is defined by:
\begin{equation}
L_{reg} = \sum_{k\in \{x,y,z,r\}}S(d_k,\hat{d_k})，
\end{equation}
where the loss metric is a smoothed L1-norm function:
\begin{equation*}
S(d,\hat{d}) = \left\{
                \begin{array}{ll}
                |d-\hat{d}|, & \text{if}~ |d_k-\hat{d}|>1, \\
                (d-\hat{d})^2, & \text{else}.
                \end{array}
              \right.\\
\end{equation*}

The loss function for each anchor box is defined by:
\begin{equation}
L = L_{cls}+ pL_{reg}.
\end{equation}
This equation indicates that the regression loss only applies to positive samples because only in these cases $p=1$.
The overall loss function is the mean of loss function for some selected anchor boxes.
We use positive sample balancing and hard negative mining to do the selection (see the next subsection).


\begin{figure}[htbp]
    \begin{subfigure}[t]{\columnwidth}
        \includegraphics[width=\columnwidth]{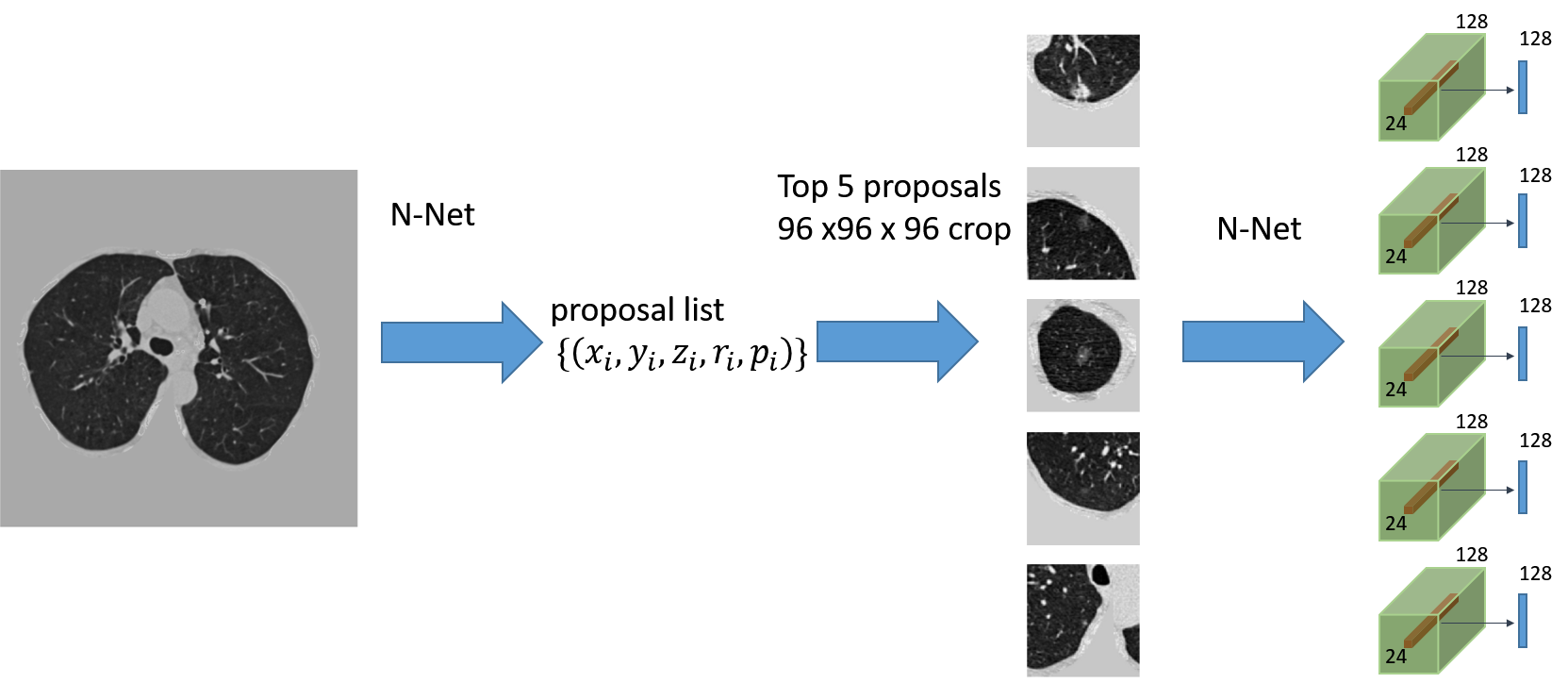}
        \caption{}
        \label{pre_casenet}
    \end{subfigure}
    \begin{subfigure}[t]{\columnwidth}
    \includegraphics[width=\columnwidth]{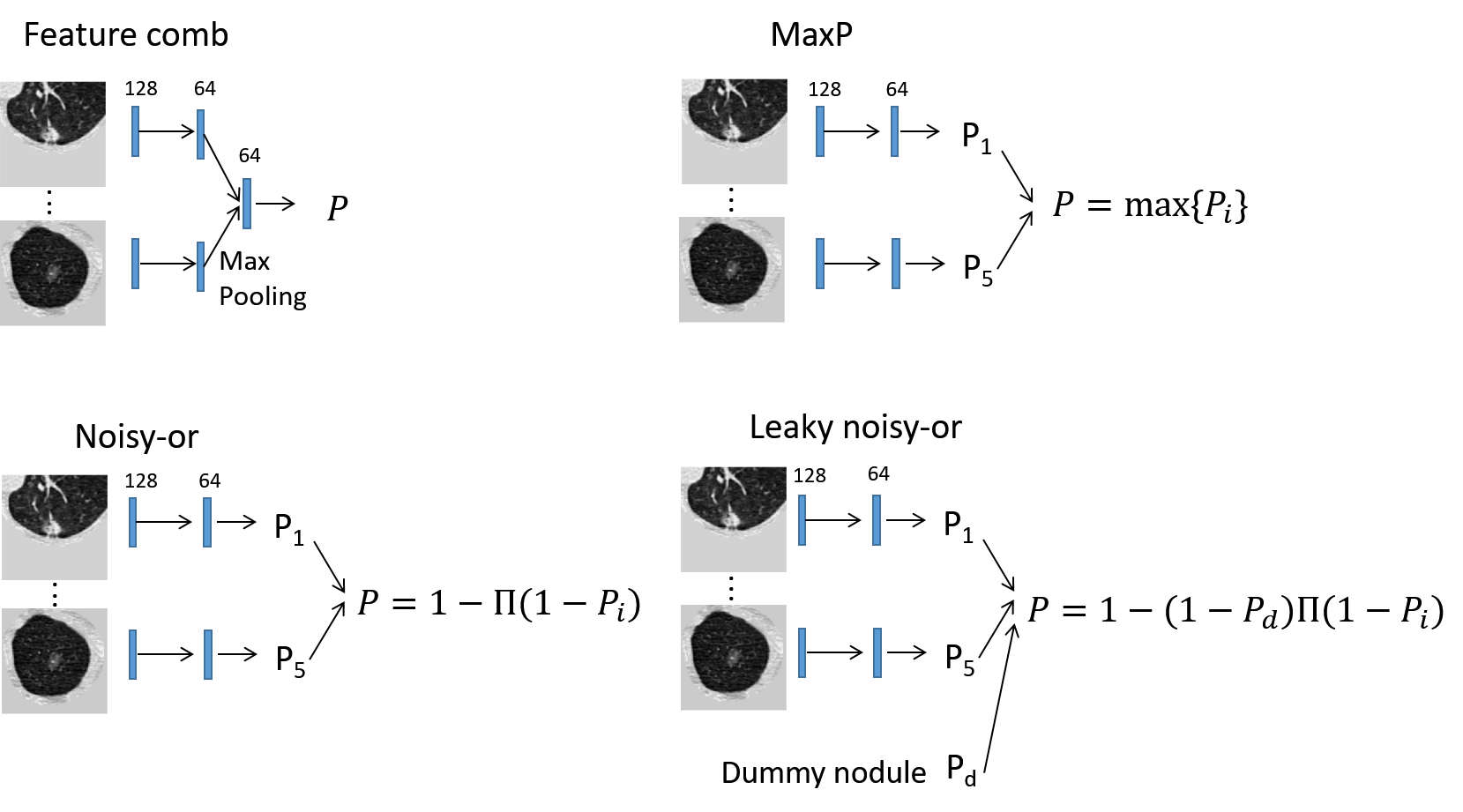}
    \caption{}
    \label{fig_nodulecomb}
    \end{subfigure}

    \caption{Illustration of the case classifier. (a) The procedures of getting proposals and features of proposals. (b) Different multi-nodule information integration methods.}
\end{figure}

\subsection{Positive sample balancing}
For a big nodule, there are many corresponding positive anchor boxes. To reduce the correlation among training samples, only one of them is randomly chosen in the training phase.

Though we have removed some very small nodules from LUNA, the distribution of nodule size is still highly unbalanced. The number of small nodules is much larger than that of big nodules. If uniform sampling is used, the trained network will bias small nodules. This is unwanted because big nodules are usually stronger indicators of cancer than smaller ones. Therefore, the sampling frequencies of big nodules are increased in the training set. Specifically, the sampling frequency of nodules larger than 30 mm and 40 mm are 2 and 6 times higher than other nodules, respectively. 

\subsection{Hard negative mining}
There are much more negative samples than positive samples. Though most negative samples can be easily classified by the network, a few of them have similar appearances with nodules and are hard to be classified correctly. A common technique in object detection, hard negative mining is used to deal with this problem. We use a simple online version of hard negative mining in training.

First, by inputting the patches to the network, we obtain the output map, which stands for a set of proposed bounding boxes with different confidences. Second, $N$ negative samples are randomly chosen to form a candidate pool. Third, the negative samples in this pool are sorted in descending order based on their classification confidence scores, and the top $n$ samples are selected as the hard negatives. Other negative samples are discarded and not included in the computation of loss. The use of a randomly selected candidate pool can reduce the correlation between negative samples. By adjusting the size of the candidate pool and the value of $n$, the strength of hard negative mining can be controlled.

\subsection{Image splitting during testing}
After the network is trained, the entire lung scans could be used as input to obtain all suspicious nodules. Because the network is fully convolutional, it is straightforward to do this. But it is infeasible with our GPU memory constraint. Even though the network needs much fewer memory in testing than in training, the requirement still exceeds the maximum memory of the GPU. To overcome this problem, we split the lung scans into several parts ($208\times208\times208\times1$ per part), process them separately, and then combine the results. We keep these splits overlapped by a large margin (32 pixels) to eliminate the unwanted border effects during convolution computations.

This step will output many nodule proposals $\{x_i, y_i, z_i,r_i, p_i \}$ where $x_i, y_i, z_i$ stand for the center of the proposal, $r_i$ stands for the radius, and $p_i$ stands for the confidence. Then a non-maximum suppression (NMS) \cite{girshick_rich_2014} operation is performed to rule out the overlapping proposals. Based on these proposals, another model is used to predict cancer probability.

\section{Cancer classification}\label{sec:classification}

Then we evaluate the cancer probability of the subject based on the nodules detected.
For each subject, five proposals are picked out based on their confidence scores in N-Net. As a simple way of data augmentation, during training, proposals are picked stochastically. The probability of being picked for a nodule is proportional to its confidence score. But during testing, top five proposals are directly picked. If the number of detected proposals is smaller than five, several blank images are used as inputs so that the number is still five.

Due to the limited number of training samples, it is unwise to build an independent neural network to do this, otherwise over-fitting will occur. An alternative is to re-use the N-Net trained in the detection phase. 

For each selected proposal,  we crop a $96\times 96\times96 \times 1$ patch whose center is the nodule (notice that this patch is smaller than that in the detection phase),  feed it to the N-Net, and get the last convolutional layer of N-Net, which has a size of $24 \times 24 \times 24 \times 128$. The central $2 \times 2 \times 2$ voxels of each proposal are extracted and max-pooled, resulting in a 128-D feature (Fig. \ref{pre_casenet}). To get a single score from multiple nodules for a single case, four integration methods are explored (See Fig. \ref{fig_nodulecomb}).

\subsection{Feature combining method}
First, the features of all top five nodules are fed to a fully connected layer to give five 64-D features. These features are then combined to give a single 64-D feature by max-pooling. The feature vector is then fed to the second fully connected layer, whose activation function is the sigmoid function, to get the cancer probability of the case (Left panel in Fig. \ref{fig_nodulecomb}).

This method may be useful if there exists some nonlinear interaction between nodules. A disadvantage is that it lacks interpretability in the integration step as there is no direct relationship between each nodule and the cancer probability.

\subsection{MaxP method}
The features of all top five nodules are separately fed into the same two-layer Perceptron with 64 hidden unit and one output unit. The activation function of the last layer is also the sigmoid function, which outputs the cancer probability of every nodule. Then the maximum of those probabilities is taken as the probability of the case.

Compared with the feature combining method, this method provides interpretability for each nodule. Yet this method neglects the interaction between nodules. For example, if a patient has two nodules which both have 50\% cancer probability, the doctors would infer that the overall cancer probability is much larger than 50\%, but the model would still give a prediction of 50\%.

\subsection{Noisy-or method}
To overcome the problem mentioned above, we assume that the nodules are independent causes of cancer, and the malignancy of any one leads to cancer.  Like the maximal probability model, the feature of every nodule is first fed to a two-layer Perceptron to get the probability. The final cancer probability is \cite{pearl2014probabilistic}:
\begin{equation}
    P = 1 - \prod_i (1 - P_i),
\end{equation}
where $P_i$ stands for the cancer probability of the $i$-th nodule.

\subsection{Leaky Noisy-or method}
There is a problem in the Noisy-or method and MaxP method. If a subject has cancer but some malignant nodules are missed by the detection network, these methods would attribute the cause of cancer to those detected but benign nodules, which would increase the probabilities of other similar benign nodules in the dataset. Clearly, this does not make sense.
We introduce a hypothetical dummy nodule, and define $P_d$ as its cancer probability \cite{pearl2014probabilistic}. The final cancer probability becomes:
\begin{equation}
    P = 1 - (1 - P_d)\prod_i (1 - P_i).
\end{equation}
$P_d$ is learned automatically in the training procedure instead of manually tuned.

This model is used as our default model, which is called C-Net (C stands for case).


\subsection{Training procedure}
\label{sec_training}
The standard cross-entropy loss function is used for case classification.
Due to the memory constraint, the bounding boxes for nodules of each case are generated in advance. The classifier, including the shared feature extraction layers (the N-Net part) and integration layers, is then trained over these pre-generated bounding boxes.
Since the N-Net is deep, and the 3D convolution kernels have more parameters than 2D convolution kernels, yet the number of samples for classification is limited, the model tends to over-fit the training data.

To deal with this problem, two methods are adopted: data augmentation and alternate training. 3D data-augmentation is more powerful than 2D data augmentation. For example, if we only consider flip and axis swap, there are 8 variants in the 2D case, but 48 variants in the 3D case.
Specifically, the following data augmentation methods are used: (1) randomly flipping in 3 directions (2) resizing by a random number between 0.75 and 1.25, (3) rotating by any angle in 3D, (4) shifting in 3 directions with a random distance smaller than 15\% of the radius. 
Another commonly used method to alleviate over-fitting is to use some proper regulizers. In this task, since the convolutional layers are shared by the detector and classifier, these two tasks can naturally be regulizers of each other. So we alternately train the model on the detector and classifier. Specifically, in each training block, there is a detector training epoch and a classifier training epoch. 

The training procedure is quite unstable because the batch size is only 2 per GPU, and there are many outliers in the training set. Gradient clipping is therefore used in a later stage of training, i.e. if the $l_2$ norm of the gradient vector is larger than one, it would be normalized to one.

Batch normalization (BN) \cite{ioffe2015batch} is used in the network. But directly applying it during alternate training is problematic. In the training phase, the BN statistics (average and variance of activation) are calculated inside the batch, and in the testing phase, the stored statistics (the running average statistics) are used. The alternate training scheme would make the running average unsuitable for both classifier and detector. First, the input samples of them are different: the patch size is 96 for the classifier and 128 for the detector. Second, the center of the patch is always a proposal for the classifier, but the image is randomly cropped for the detector. So the average statistics would be different for these two tasks, and the running average statistics might be at a middle point and deteriorate the performance in the validation phases for both of them.
To solve this problem, we first train the classifier, making the BN parameters suitable for classification. Then at the alternate training stage, these parameters are frozen, i.e. during both the training and validation phases, we use the stored BN parameters.

In summary, the training procedure has three stages: (1) transfer the weights from the trained detector and train the classifier in the standard mode, (2) train the classifier with gradient clipping, then freeze the BN parameters, (3) train the network for classification and detection alternately with gradient clipping and the stored BN parameters. This training scheme corresponds to $A\rightarrow B \rightarrow E$ in Table \ref{tab_training}.

\section{Results}\label{sec:results}

\subsection{Nodule detection}
\begin{figure}
\centering
\begin{subfigure}[t]{0.75\columnwidth}
\includegraphics[width=\columnwidth]{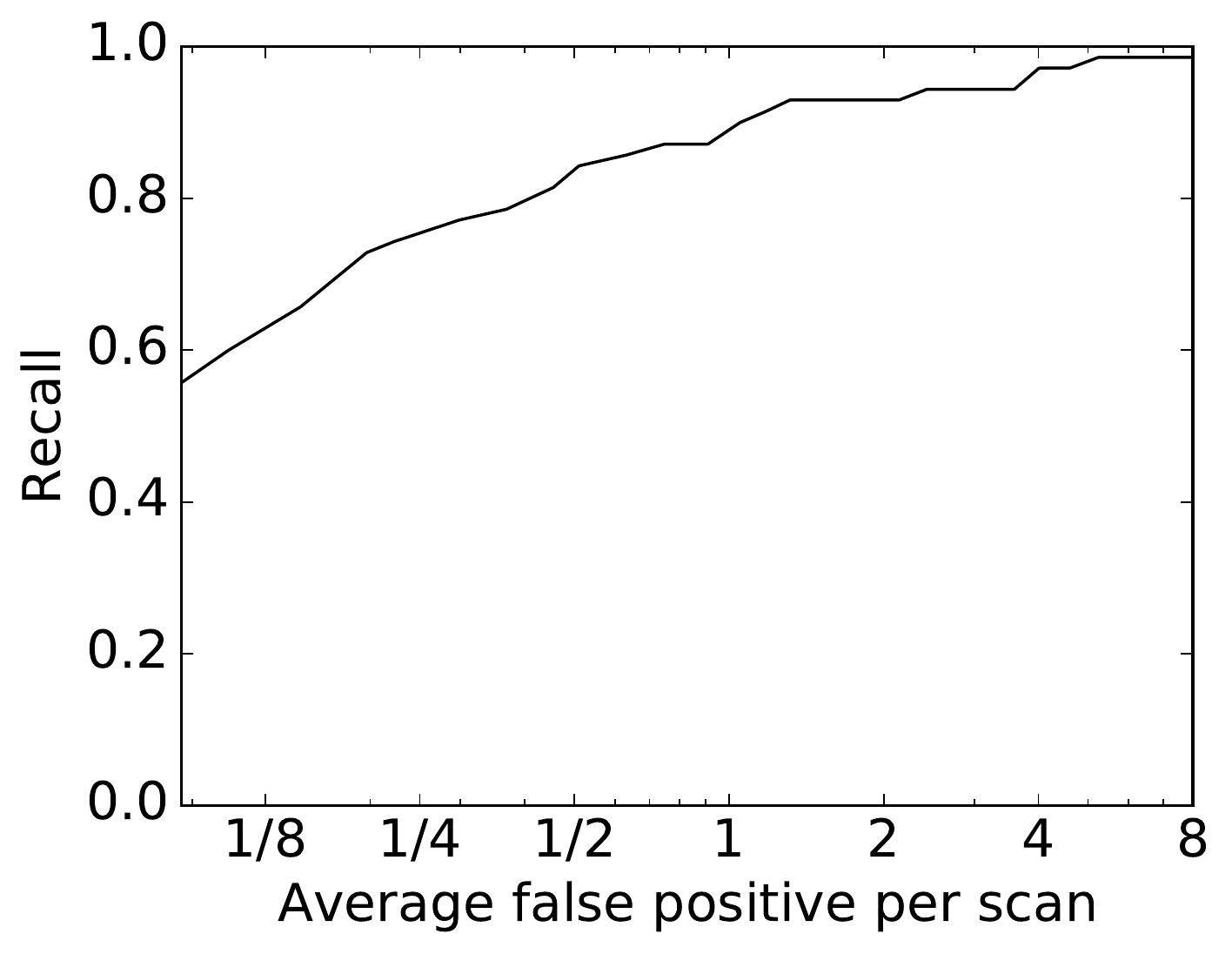}
\subcaption{}
\label{fig_froc}
\end{subfigure}
\begin{subfigure}[t]{0.75\columnwidth}
\includegraphics[width=\columnwidth]{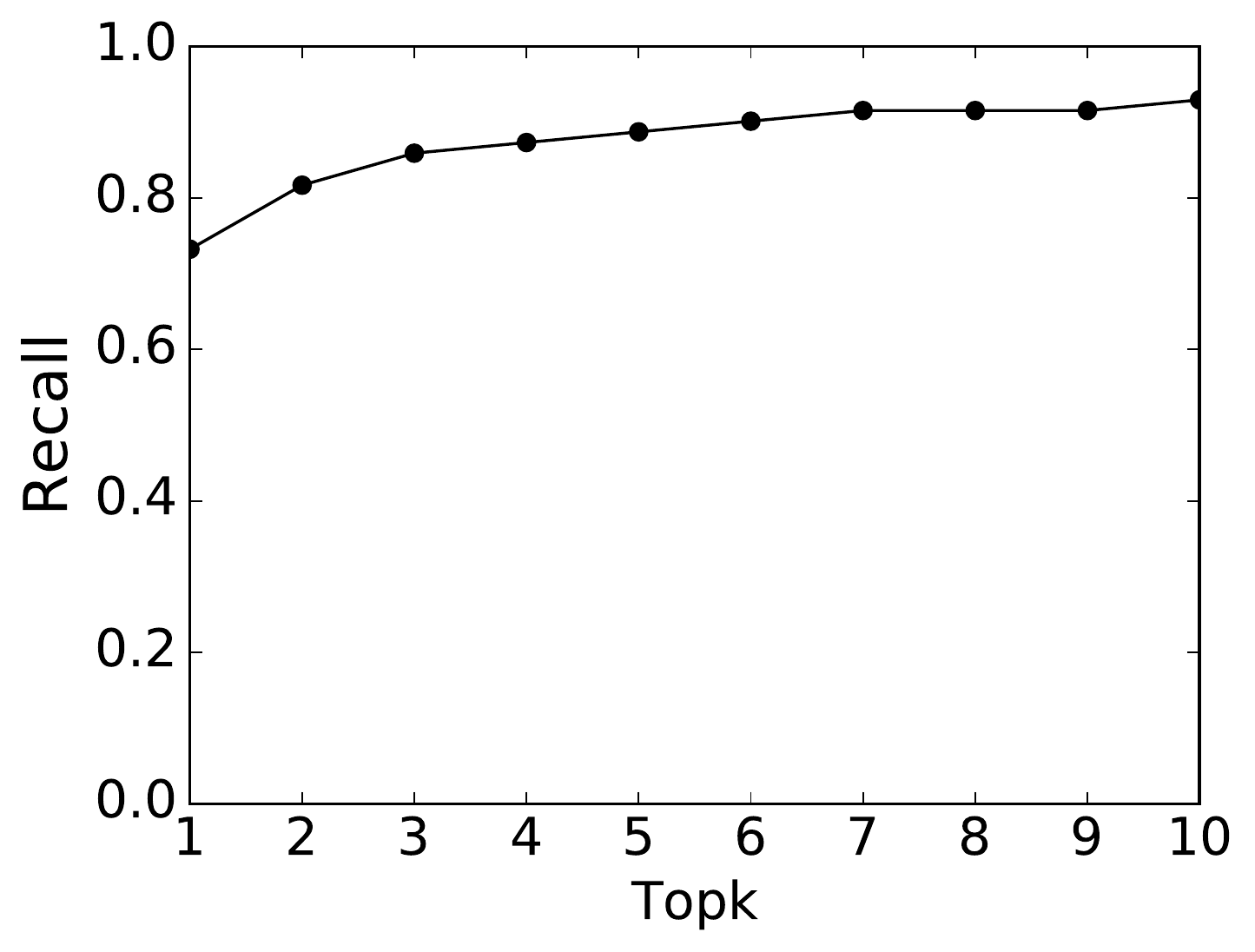}
\subcaption{}
\label{fig_topk}
\end{subfigure}
\caption{The result of detection module. (a) The FROC curve. (b) the recall at different topk level.}
\end{figure}

Because our detection module is designed to neglect the very small nodules during training, the LUNA16 evaluation system is not suitable for evaluating its performance. We evaluated the performance on the validation set of DSB. It contains data from 198 cases and there are 71 (7 nodules smaller than 6 mm are ruled out) nodules in total. The Free Response Operating Characteristic (FROC) curve is shown in Fig. \ref{fig_froc}. The average recall at 1/8, 1/4, 1/2, 1, 2, 4, 8 false positive per scan is 0.8562. 

We also investigated the recall when a different top-k number was chosen (Fig. \ref{fig_topk}). The result showed that $k=5$ was enough to capture most nodules. 

\subsection{Case classification}

\begin{table}[]
    \centering
    \caption{The cross-entropy loss of different training methods on the test set. The third block shows the performance of top 4 teams in the competition.}
    \label{tab_training}
\begin{threeparttable}
    \begin{tabular}{@{}ll@{}}
        \toprule
        Training method                                                                                        & Loss   \\ \midrule
        C-Net                                                                                     & 1.2633 \\
        (A) C-Net + Aug                                                                                             & 0.4173 \\
        (B) C-Net +Aug + Clip                                                                                      & 0.4157 \\
        (C) C-Net +Aug + Alt                                                                        & 0.4060 \\
        (D) C-Net +Aug + Alt + Clip                                                                 & 0.4185 \\
        \begin{tabular}[c]{@{}l@{}}
        (E) C-Net +Aug + Alt + Clip +\\ BN freeze \end{tabular} & 0.412  \\ \hline \hline
        A $\rightarrow$ B                                                                                                    & 0.4060 \\
        A $\rightarrow$ B $\rightarrow$ D                                                                                                  & 0.4024 \\
        A $\rightarrow$ B $\rightarrow$ E                                                                                                & \textbf{0.3989} \\ \hline \hline
grt123                                                                                                 & 0.3998 \\
Julian de Wit \& Daniel Hammack                                                                        & 0.4012 \\
Aidence                                                                                                & 0.4013 \\
qfpxfd                                                                                                 & 0.4018 \\ \bottomrule
    \end{tabular}

    \begin{tablenotes}
    \item  Aug: data augmentation; Clip: gradient clipping; Alt: alternate training; BN freeze: freezing batch normalization parameters. \\ grt123 is the name of our team. The training scheme is slightly different from that in the competition.
    \end{tablenotes}

\end{threeparttable}
\end{table}
To select the training schemes, we rearranged the training set and the validation set because we empirically found that the original training and validation set differed significantly. One-fourth of the original training set was used as the new validation set. The rest were combined with the original validation set to form the new training set.

As described in Section \ref{sec_training}, four techniques were used during training: (1) data augmentation, (2) gradient clipping, (3) alternate training, (4) freezing BN parameters. Different combinations of these techniques (denoted by A, B... E in Table \ref{tab_training}) and different orders of stages were explored on the new validation set. It was found that the $A\rightarrow B \rightarrow E$ scheme performed the best. 
After the competition, we could still submit results to the evaluation server, so we evaluated the training schemes on the test set. Table \ref{tab_training} shows the results on the test set (the models were trained on the union of the training set and the validation set). It was found that $A\rightarrow B \rightarrow E$ was indeed the best one among many schemes.

From block 1 in Table \ref{tab_training} we can draw several conclusions. First, without data augmentation, the model would seriously over-fit the training set. Second, alternate training improved the performance significantly. Third, gradient clipping and BN freezing were not very useful in these schemes.

From block 2 in Table \ref{tab_training}, it is found that clipping was useful when we finetuned the result of stage A (A$\rightarrow$B). And the alternative training was useful to further finetune the model (A$\rightarrow$B$\rightarrow$D). In addition, introducing the BN freezing technique further improved the result (A$\rightarrow$B$\rightarrow$E).

The block 3 in Table \ref{tab_training} shows the performance of top 4 teams in the competition. The scores are very close. But we achieved the highest score with a single model. 

The results of different multi-nodule information integration models are shown in Table \ref{tab_comb}. The models were all trained using the alternate training method (configuration C in Table \ref{tab_training}). The three probability-based methods were much better than the feature combining method. And the Leaky Noisy-or model performed the best.

\begin{table}
    \centering
    \caption{The cross-entropy loss of different nodule combining methods on the test set.}
    \label{tab_comb}
    \begin{tabular}{@{}ll@{}}
    
        \toprule
        \begin{tabular}[c]{@{}l@{}}Name \end{tabular} & Loss   \\ \midrule
        Feature comb                                    & 0.4286 \\
        MaxP                                        & 0.4090  \\
        Noisy-or                                        & 0.4185  \\
        Leaky noisy-or                                     & \textbf{0.4060}  \\ \bottomrule
    \end{tabular}
\end{table}

\begin{figure}
\centering
    \begin{subfigure}[t]{.49\columnwidth}
        \centering
        \includegraphics[width=\columnwidth]{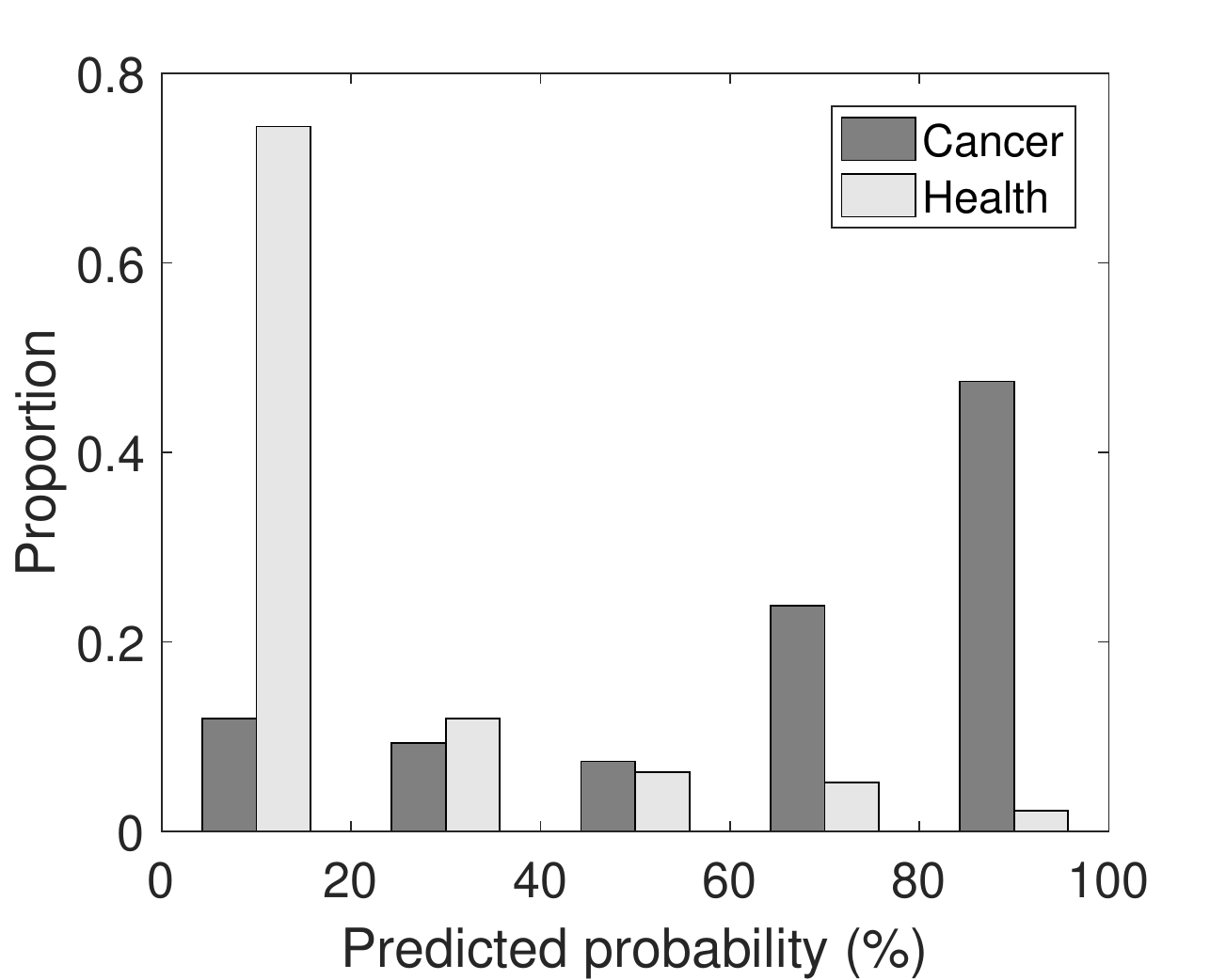}
        \caption{}
        \label{fig_hist_train}
    \end{subfigure}
    \begin{subfigure}[t]{.49\columnwidth}
        \centering
        \includegraphics[width=\columnwidth]{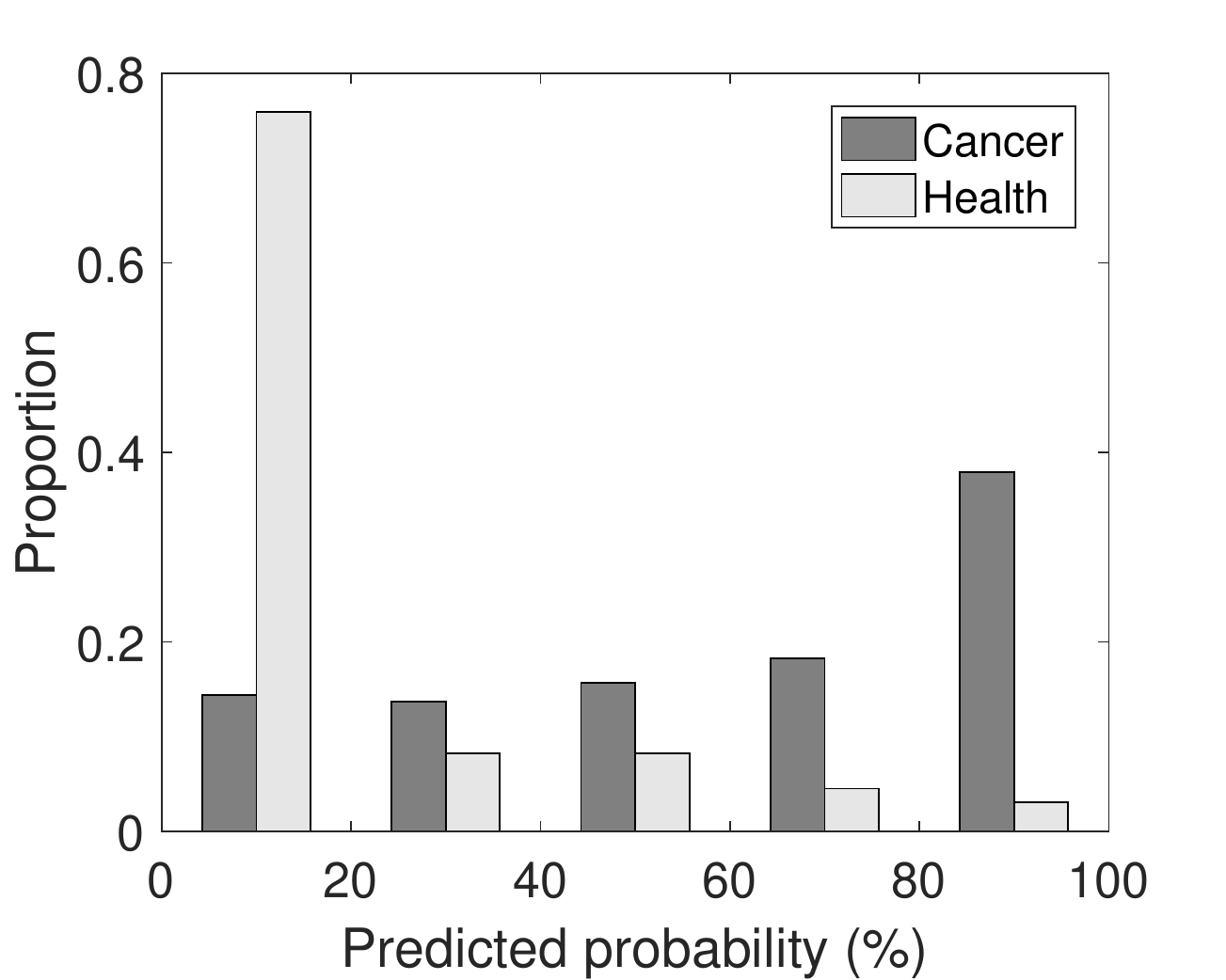}
        \caption{}
        \label{fig_hist_test}
    \end{subfigure}%
    
    \begin{subfigure}[t]{.49\columnwidth}
        \centering
        \includegraphics[width=\columnwidth]{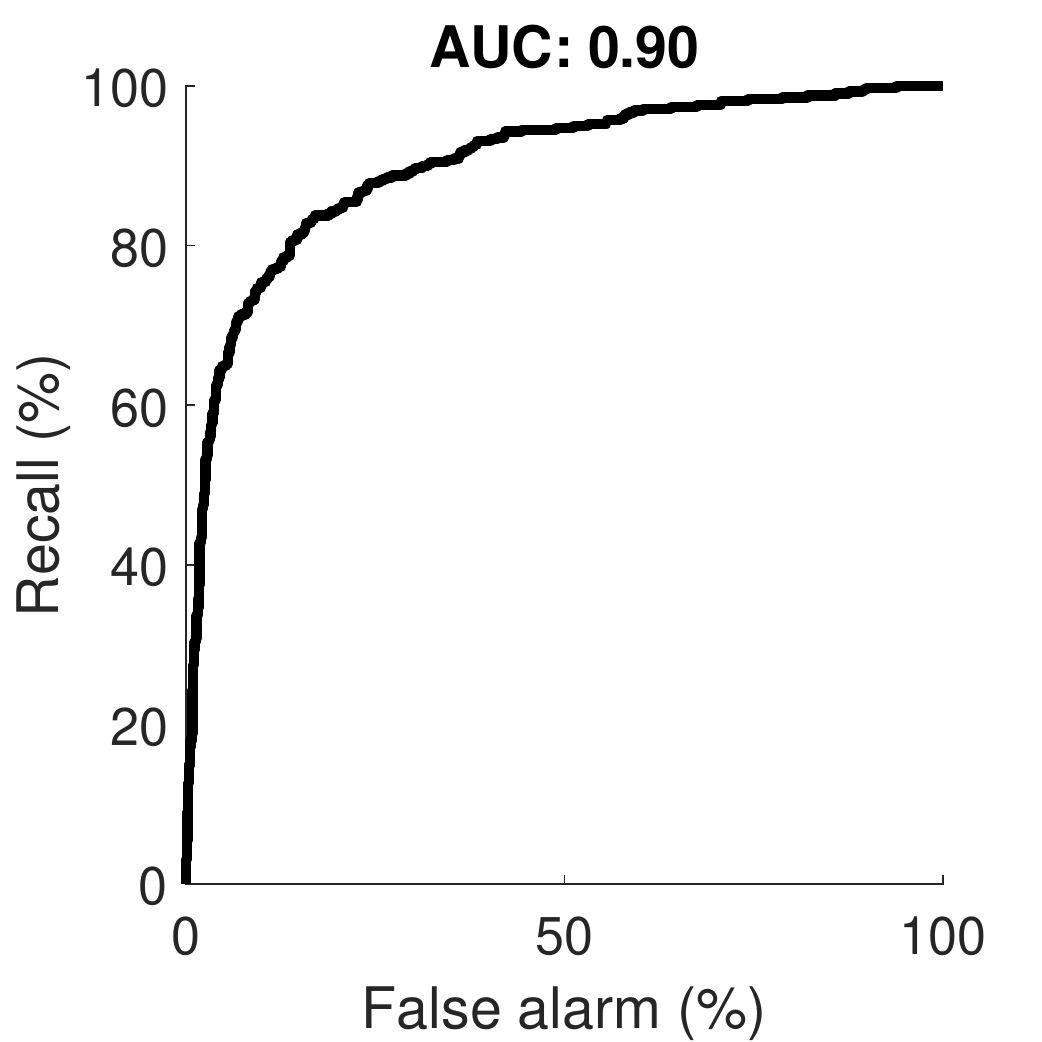}
        \caption{}
        \label{fig_roc_train}
    \end{subfigure}
    \begin{subfigure}[t]{.49\columnwidth}
        \centering
        \includegraphics[width=\columnwidth]{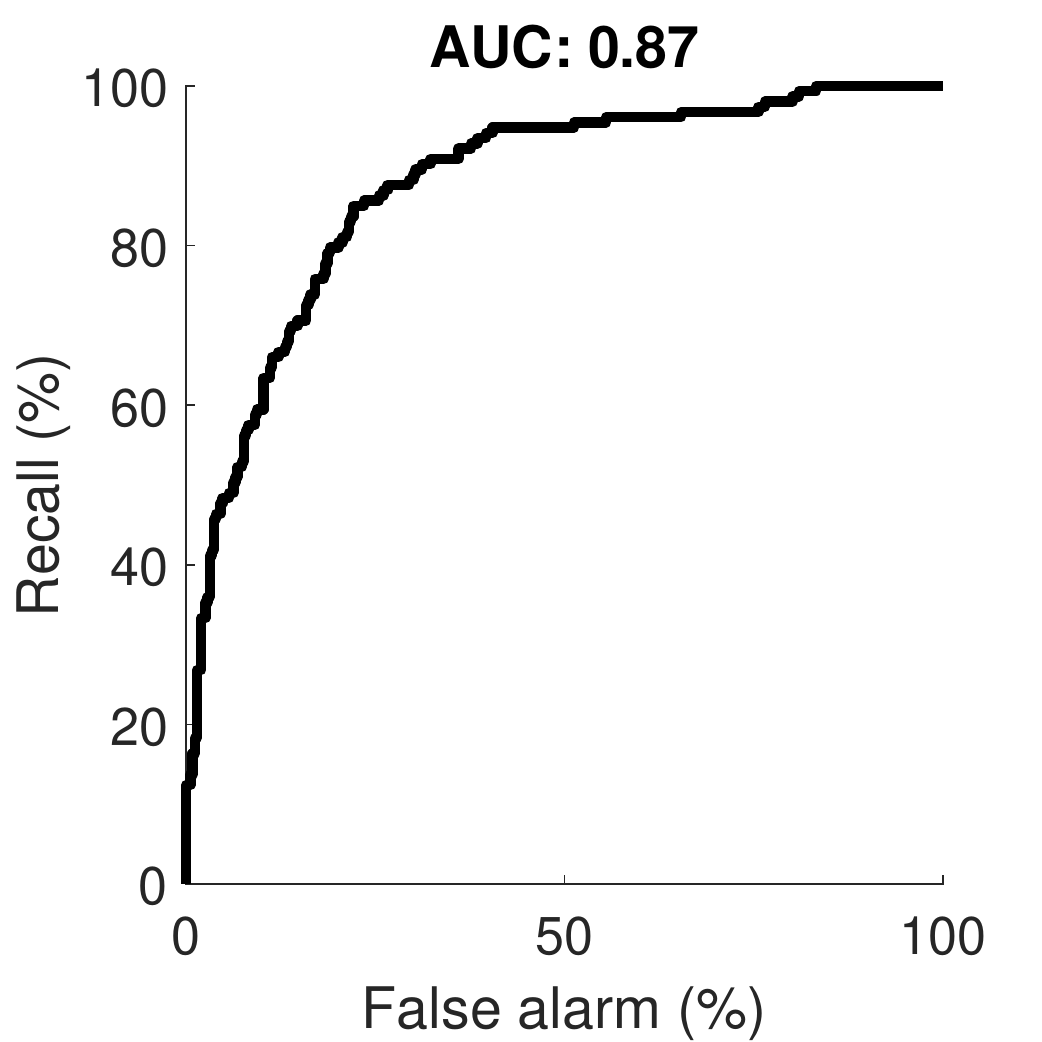}
        \caption{}
        \label{fig_roc_test}
    \end{subfigure}

    \caption{Cancer classification results. The distribution of predicted cancer probability for healthy people and patients on the training (a) and test set (b). The ROC curve of cancer classification task on the training (c) and test set (D).}
    \label{fig_roc}
\end{figure}

The distributions of the predicted cancer probability on the training and test sets are shown in Fig. \ref{fig_roc}a,b. A Receiver Operating Characteristic (ROC) curve was obtained on each set by varying the threshold (Fig. \ref{fig_roc}c,d). The areas under the ROC curves (AUC) were 0.90 and 0.87 on the training and test set, respectively. If we set the threshold to 0.5 (classified as cancer if the predicted probability is higher than the threshold), the classification accuracies were 85.96\% and 81.42\% on the training and test sets, respectively. If we set the threshold to 1 (all cases are predicted healthy), the classification accuracies were 73.73\% and 69.76\% on the training and test sets, respectively.

\begin{figure}
    \begin{subfigure}[b]{\columnwidth}
        \includegraphics[width=\columnwidth]{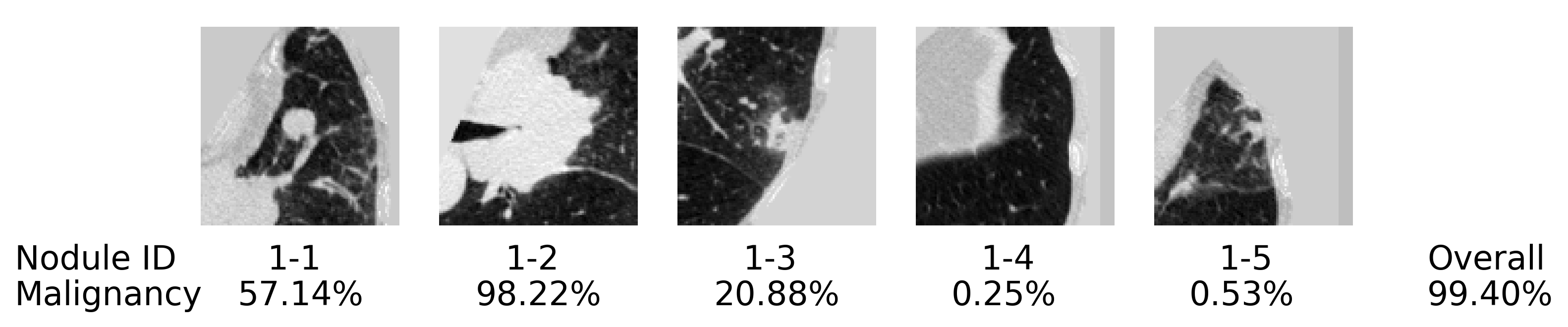}
    \end{subfigure}
    \begin{subfigure}[b]{\columnwidth}
        \includegraphics[width=\columnwidth]{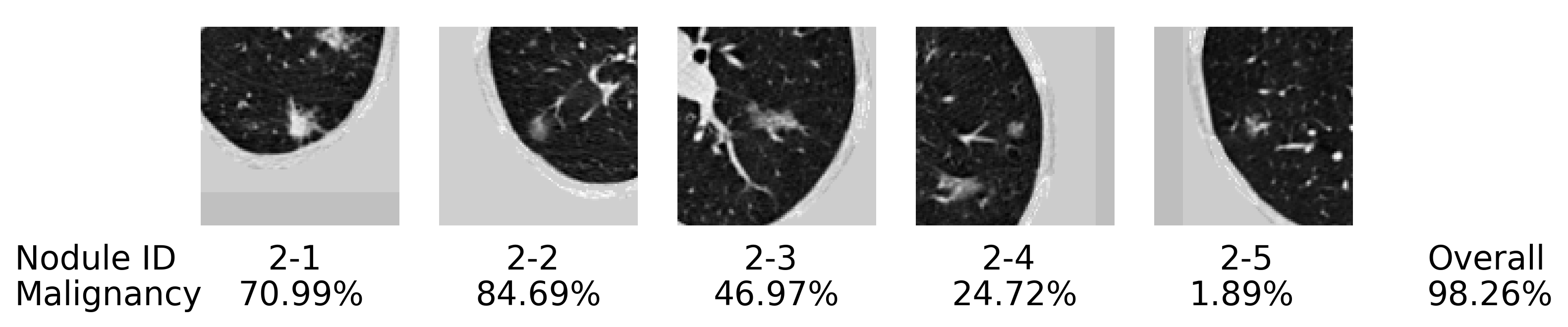}
        \subcaption{}
    \end{subfigure}
    
    \begin{subfigure}[b]{\columnwidth}
        \includegraphics[width=\columnwidth]{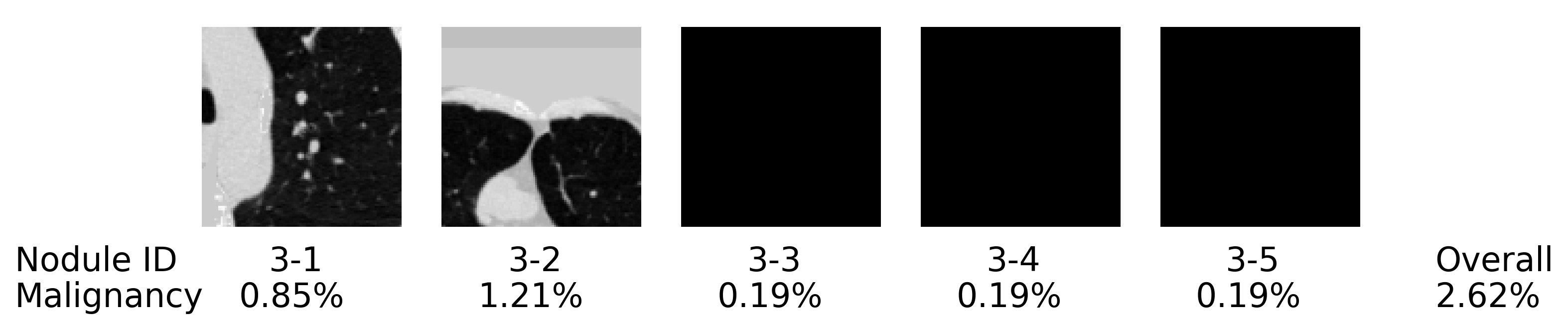}
    \end{subfigure}
    \begin{subfigure}[b]{\columnwidth}
        \includegraphics[width=\columnwidth]{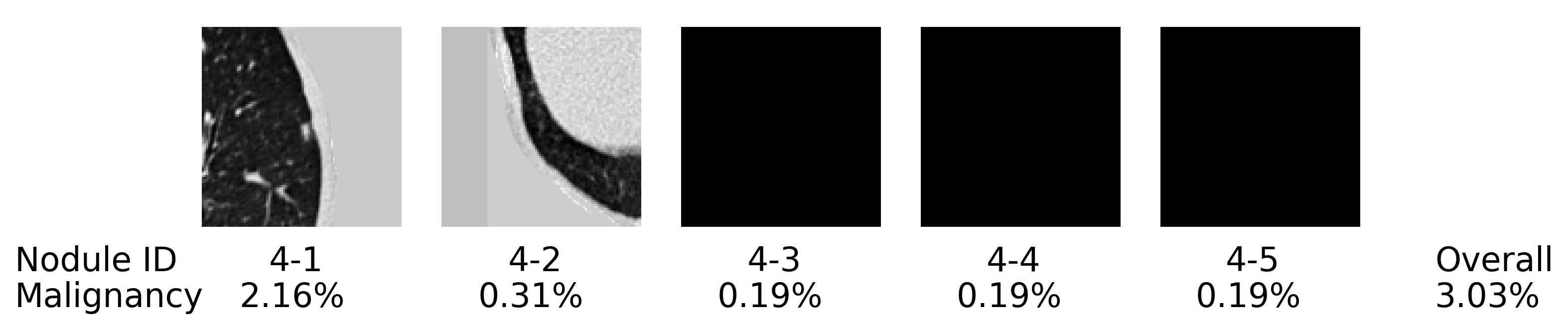}
        \subcaption{}
    \end{subfigure}
    
    \begin{subfigure}[b]{\columnwidth}
        \includegraphics[width=\columnwidth]{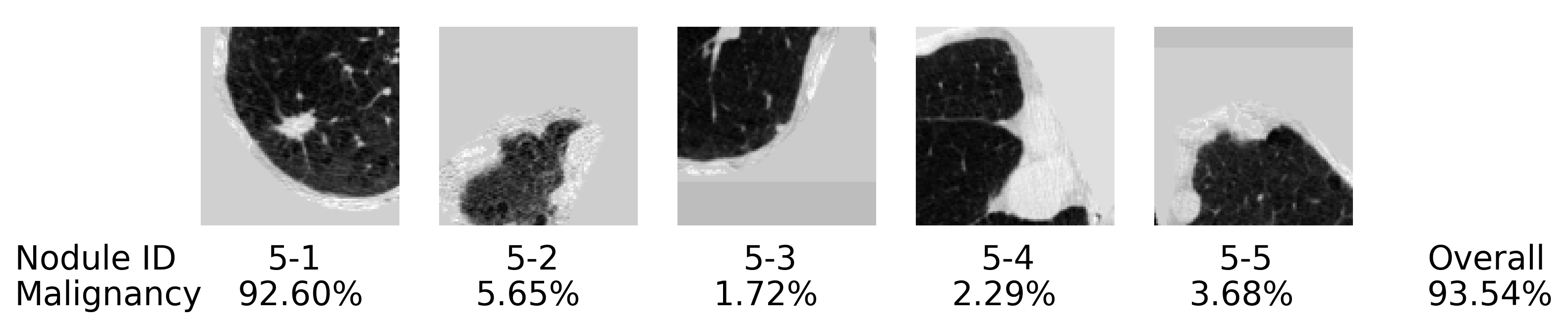}
    \end{subfigure}
    \begin{subfigure}[b]{\columnwidth}
        \includegraphics[width=\columnwidth]{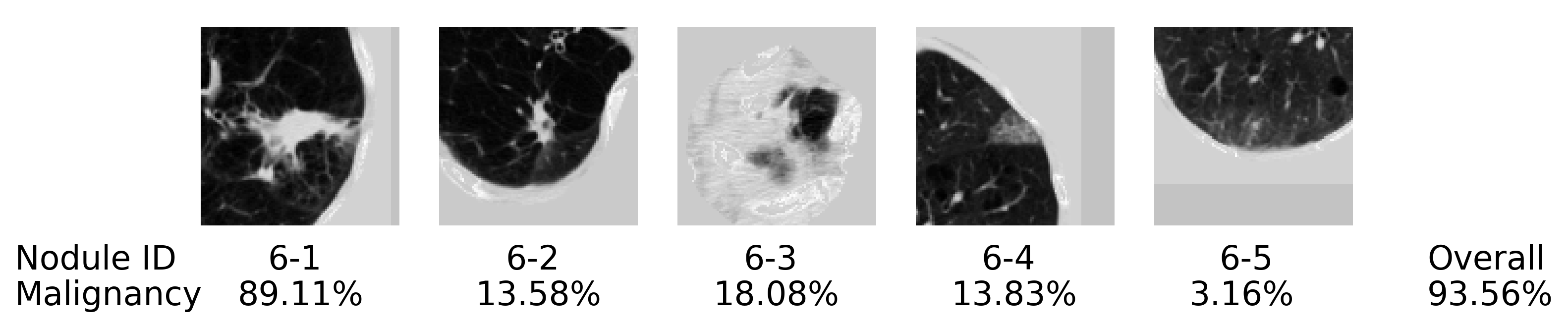}
        \subcaption{}
    \end{subfigure}
    
    \begin{subfigure}[b]{\columnwidth}
        \includegraphics[width=\columnwidth]{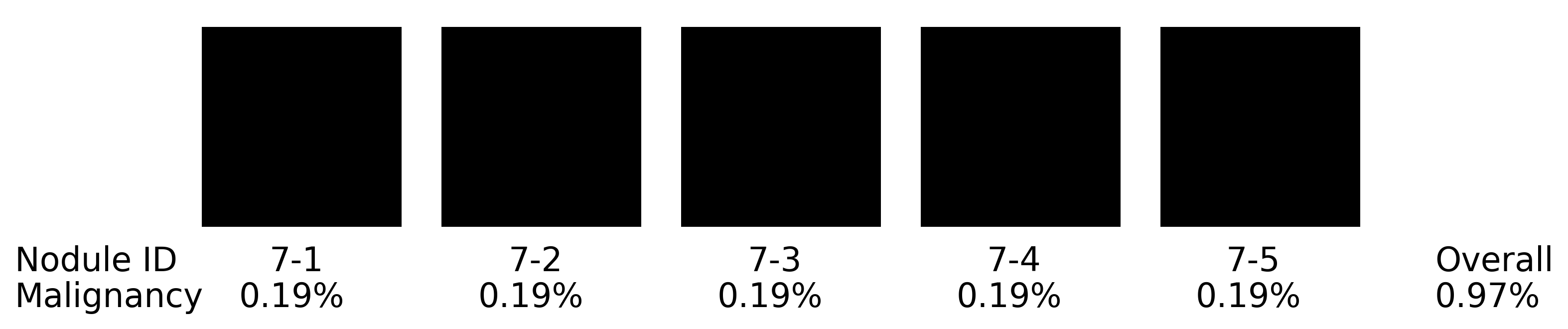}
    \end{subfigure}
    \begin{subfigure}[b]{\columnwidth}
        \includegraphics[width=\columnwidth]{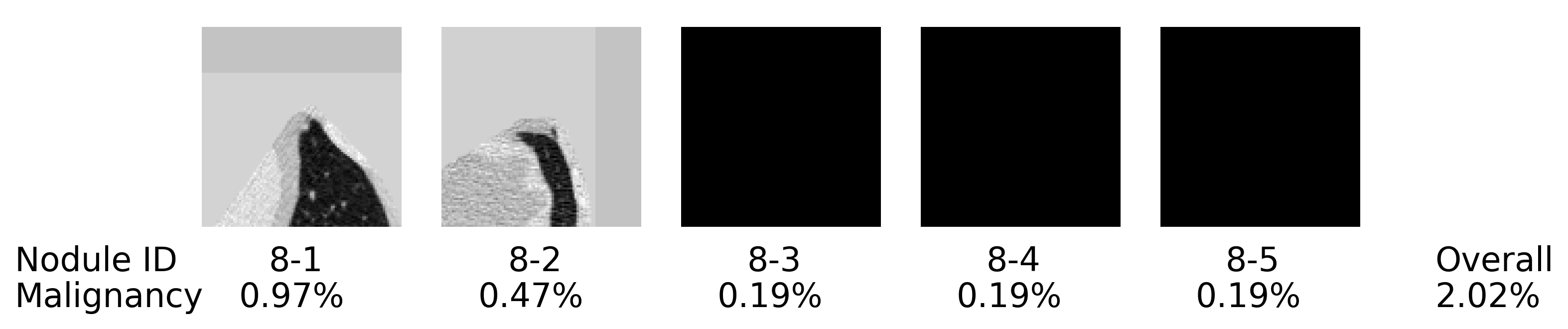}
        \subcaption{}
    \end{subfigure}
    
    \caption{The model output of several cases. (a) True positive samples. (b) False negative samples. (c) False positive samples. (d) True negative samples. When the number of detected nodules is less than five, several blank images are used as inputs.}
    \label{fig_casedemo}
\end{figure}

The classification results of several cases are shown in Fig. \ref{fig_casedemo}. For the two true positive cases (Cases 1 and 2), the model correctly predicted high cancer probabilities for both of them. Case 1 had a very large tumor (nodule 1-2), which contributed a very high cancer probability. Case 2 had several middle-sized nodules, three of which contributed significant cancer probability so that the overall probability was very high. In addition, the model learned to judge malignancy based on not only size but also the morphology. Nodule 1-1 had a larger size than nodules 2-1 and 2-2, but had a lower cancer probability. The reason is as follows. Nodule 1-1 had solid luminance, round shape, and clear border, which are indications of benignancy. While nodule 2-1 had an irregular shape and unclear border, and nodule 2-2 had opaque luminance, which are all indications of malignancy. Nodule 2-1 is called spiculated nodule, and nodule 2-2 is called part-solid ground-glass nodule \cite{ha2014pulmonary}, both of which are highly dangerous nodules. There was no significant nodule in the two false negative cases (Case 3 and 4), so their overall probability was very low. Both of the false positive cases (Case 5 and 6) had highly suspicious nodules, making them hard to be correctly classified. No nodule was detected in Case 7 and only two insignificant nodules were detected in Case 8, so the two cases were predicted as healthy, which are correct.

\section{Discussion}\label{sec:discussion}

A neural network-based method is proposed to perform automatic lung cancer diagnosis. A 3D CNN is designed to detect the nodules and a leaky noisy-or model is used to evaluate the cancer probability of each detected nodule and combine them together. 
The overall system achieved very good results on the cancer classification task in a benchmark competition.


The proposed leaky noisy-or network may find many applications in medical image analysis. Many disease diagnosis starts with an image scanning. The lesion(s) shown in the image may relate to the disease but the relationship is uncertain, the same situation as in the cancer prediction problem studied in this work. The leaky noisy-or model can be used to integrate the information from different lesions to predict the result. It also alleviates the demand for highly accurate fine-scaled labels.

Applying 3D CNN to 3D object detection and classification faces two difficulties. First, the model occupies much more memory when the model size grows, so the running speed, batch size, and model depth are all limited. We designed a shallower network and used image patches instead of the whole image as the input. Second, the number of parameters of a 3D CNN is significantly larger than that of a 2D CNN with similar architecture, thus the model tends to over-fit the training data. Data augmentation and alternate training are used to mitigate the problem. 


There are some potential ways to improve the performance of the proposed model. The most straightforward way is increasing the number of training samples: 1700 cases are too few to cover all variations of the nodules, and an experienced doctor sees much more cases in his career. 
Second, incorporating the segmentation labels of nodules may be useful because it has been shown that the co-training of segmentation and detection tasks can improve the performance of both tasks \cite{he_mask_2017}.

Though many teams have achieved good results in this cancer prediction competition, this task itself has an obvious limitation for the clinic: the growing speed of nodules is not considered. In fact, fast-growing nodules are usually dangerous. To detect the growing speed, one needs to scan the patient multiple times during a period and detect all nodules (not only large nodules but also small nodules) and align them along time. Although the proposed method in this work does not pursue high detection accuracy for small nodules, it is possible to modify it for this purpose. For example, one can add another unpooling layer to incorporate finer-scale information and reduce the anchor size.

\bibliographystyle{IEEEtranN}
\bibliography{IEEEabrv,kaggle3}

\begin{thebibliography}{42}
\providecommand{\natexlab}[1]{#1}
\providecommand{\url}[1]{#1}
\csname url@samestyle\endcsname
\providecommand{\newblock}{\relax}
\providecommand{\bibinfo}[2]{#2}
\providecommand{\BIBentrySTDinterwordspacing}{\spaceskip=0pt\relax}
\providecommand{\BIBentryALTinterwordstretchfactor}{4}
\providecommand{\BIBentryALTinterwordspacing}{\spaceskip=\fontdimen2\font plus
\BIBentryALTinterwordstretchfactor\fontdimen3\font minus
  \fontdimen4\font\relax}
\providecommand{\BIBforeignlanguage}[2]{{%
\expandafter\ifx\csname l@#1\endcsname\relax
\typeout{** WARNING: IEEEtranN.bst: No hyphenation pattern has been}%
\typeout{** loaded for the language `#1'. Using the pattern for}%
\typeout{** the default language instead.}%
\else
\language=\csname l@#1\endcsname
\fi
#2}}
\providecommand{\BIBdecl}{\relax}
\BIBdecl
\renewcommand{\BIBentryALTinterwordstretchfactor}{4}

\bibitem[Infante et~al.(2009)Infante, Cavuto, Lutman, Brambilla, Chiesa,
  Ceresoli, Passera, Angeli, Chiarenza, Aranzulla,
  et~al.]{infante_randomized_2009}
M.~Infante, S.~Cavuto, F.~R. Lutman, G.~Brambilla, G.~Chiesa, G.~Ceresoli,
  E.~Passera \emph{et~al.}, ``A randomized study of lung cancer screening with
  spiral computed tomography: three-year results from the dante trial,''
  \emph{American Journal of Respiratory and Critical Care Medicine}, vol. 180,
  no.~5, pp. 445--453, 2009.

\bibitem[Singh et~al.(2012)Singh, Gierada, Pinsky, Sanders, Fineberg, Sun,
  Lynch, and Nath]{singh2012reader}
S.~Singh, D.~S. Gierada, P.~Pinsky, C.~Sanders, N.~Fineberg, Y.~Sun, D.~Lynch
  \emph{et~al.}, ``Reader variability in identifying pulmonary nodules on chest
  radiographs from the national lung screening trial,'' \emph{Journal of
  Thoracic Imaging}, vol.~27, no.~4, p. 249, 2012.

\bibitem[Shin et~al.(2013)Shin, Orton, Collins, Doran, and
  Leach]{shin2013stacked}
H.-C. Shin, M.~R. Orton, D.~J. Collins, S.~J. Doran, and M.~O. Leach, ``Stacked
  autoencoders for unsupervised feature learning and multiple organ detection
  in a pilot study using 4d patient data,'' \emph{IEEE Transactions on Pattern
  Analysis and Machine Intelligence}, vol.~35, no.~8, pp. 1930--1943, 2013.

\bibitem[Esteva et~al.(2017)Esteva, Kuprel, Novoa, Ko, Swetter, Blau, and
  Thrun]{esteva2017dermatologist}
A.~Esteva, B.~Kuprel, R.~A. Novoa, J.~Ko, S.~M. Swetter, H.~M. Blau, and
  S.~Thrun, ``Dermatologist-level classification of skin cancer with deep
  neural networks,'' \emph{Nature}, vol. 542, no. 7639, pp. 115--118, 2017.

\bibitem[Gulshan et~al.(2016)Gulshan, Peng, Coram, Stumpe, Wu, Narayanaswamy,
  Venugopalan, Widner, Madams, Cuadros, et~al.]{gulshan_development_2016}
V.~Gulshan, L.~Peng, M.~Coram, M.~C. Stumpe, D.~Wu, A.~Narayanaswamy,
  S.~Venugopalan \emph{et~al.}, ``Development and validation of a deep learning
  algorithm for detection of diabetic retinopathy in retinal fundus
  photographs,'' \emph{The Journal of the American Medical Association}, vol.
  316, no.~22, pp. 2402--2410, 2016.

\bibitem[Litjens et~al.(2017)Litjens, Kooi, Bejnordi, Setio, Ciompi,
  Ghafoorian, van~der Laak, van Ginneken, and S{\'a}nchez]{litjens_survey_2017}
G.~Litjens, T.~Kooi, B.~E. Bejnordi, A.~A.~A. Setio, F.~Ciompi, M.~Ghafoorian,
  J.~A. van~der Laak \emph{et~al.}, ``A survey on deep learning in medical
  image analysis,'' \emph{arXiv preprint arXiv:1702.05747}, 2017.

\bibitem[Duncan and Ayache(2000)]{duncan2000medical}
J.~S. Duncan and N.~Ayache, ``Medical image analysis: Progress over two decades
  and the challenges ahead,'' \emph{IEEE Transactions on Pattern Analysis and
  Machine Intelligence}, vol.~22, no.~1, pp. 85--106, 2000.

\bibitem[Peng and Schmid(2016)]{peng2016multi}
X.~Peng and C.~Schmid, ``Multi-region two-stream {R-CNN} for action
  detection,'' in \emph{European Conference on Computer Vision}.\hskip 1em plus
  0.5em minus 0.4em\relax Springer, 2016, pp. 744--759.

\bibitem[Ding et~al.(2017)Ding, Li, Hu, and Wang]{ding_accurate_2017}
J.~Ding, A.~Li, Z.~Hu, and L.~Wang, ``\BIBforeignlanguage{en}{Accurate
  {{Pulmonary Nodule Detection}} in computed tomography images using deep
  convolutional neural networks},'' in \emph{\BIBforeignlanguage{en}{Medical
  Image Computing and Computer-Assisted Intervention 2017}}, ser. Lecture Notes
  in Computer Science.\hskip 1em plus 0.5em minus 0.4em\relax {Springer, Cham},
  Sep. 2017, pp. 559--567.

\bibitem[Armato et~al.(2009)Armato, Roberts, Kocherginsky, Aberle, Kazerooni,
  MacMahon, van Beek, Yankelevitz, McLennan, McNitt-Gray,
  et~al.]{sg_3rd_assessment_2009}
S.~G. Armato, R.~Y. Roberts, M.~Kocherginsky, D.~R. Aberle, E.~A. Kazerooni,
  H.~MacMahon, E.~J. van Beek \emph{et~al.}, ``Assessment of radiologist
  performance in the detection of lung nodules: dependence on the definition of
  “truth”,'' \emph{Academic Radiology}, vol.~16, no.~1, pp. 28--38, 2009.

\bibitem[Dietterich et~al.(1997)Dietterich, Lathrop, and
  Lozano-P{\'e}rez]{dietterich_solving_1997}
T.~G. Dietterich, R.~H. Lathrop, and T.~Lozano-P{\'e}rez, ``Solving the
  multiple instance problem with axis-parallel rectangles,'' \emph{Artificial
  Intelligence}, vol.~89, no.~1, pp. 31--71, 1997.

\bibitem[Ren et~al.(2015)Ren, He, Girshick, and Sun]{ren2015faster}
S.~Ren, K.~He, R.~Girshick, and J.~Sun, ``Faster {R-CNN}: towards real-time
  object detection with region proposal networks,'' in \emph{Advances in Neural
  Information Processing Systems}, 2015, pp. 91--99.

\bibitem[Pearl(2014)]{pearl2014probabilistic}
J.~Pearl, \emph{Probabilistic Reasoning in Intelligent Systems: Networks of
  Plausible Inference}.\hskip 1em plus 0.5em minus 0.4em\relax Morgan Kaufmann,
  2014.

\bibitem[Redmon and Farhadi(2016)]{redmon2016yolo9000}
J.~Redmon and A.~Farhadi, ``Yolo9000: better, faster, stronger,'' \emph{arXiv
  preprint arXiv:1612.08242}, 2016.

\bibitem[Liu et~al.(2016)Liu, Anguelov, Erhan, Szegedy, Reed, Fu, and
  Berg]{liu2016ssd}
W.~Liu, D.~Anguelov, D.~Erhan, C.~Szegedy, S.~Reed, C.-Y. Fu, and A.~C. Berg,
  ``{SSD}: single shot multibox detector,'' in \emph{European Conference on
  Computer Vision}.\hskip 1em plus 0.5em minus 0.4em\relax Springer, 2016, pp.
  21--37.

\bibitem[Chen et~al.(2016)Chen, Yang, Zhang, Alber, and
  Chen]{chen2016combining}
J.~Chen, L.~Yang, Y.~Zhang, M.~Alber, and D.~Z. Chen, ``Combining fully
  convolutional and recurrent neural networks for {3D} biomedical image
  segmentation,'' in \emph{Advances in Neural Information Processing Systems},
  2016, pp. 3036--3044.

\bibitem[Van~Ginneken et~al.(2010)Van~Ginneken, Armato, de~Hoop, van
  Amelsvoort-van~de Vorst, Duindam, Niemeijer, Murphy, Schilham, Retico,
  Fantacci, et~al.]{van_ginneken_comparing_2010}
B.~Van~Ginneken, S.~G. Armato, B.~de~Hoop, S.~van Amelsvoort-van~de Vorst,
  T.~Duindam, M.~Niemeijer, K.~Murphy \emph{et~al.}, ``Comparing and combining
  algorithms for computer-aided detection of pulmonary nodules in computed
  tomography scans: the anode09 study,'' \emph{Medical Image Analysis},
  vol.~14, no.~6, pp. 707--722, 2010.

\bibitem[{Samuel G., Armato III} et~al.(2015){Samuel G., Armato III}, Geoffrey,
  Luc, Michael~F., Charles~R., Anthony~P., Binsheng, Denise~R., Claudia~I.,
  Eric~A., Ella~A., Heber, Edwin~J.R., David, Alberto~M., Peyton~H.,
  Matthew~S., Roger~M., Gary~E., Daniel, Pais, Qing, Roberts, Smith, Starkey,
  Batra, Caligiuri, Farooqi, Gladish, Jude, Munden, Petkovska, Quint, Schwartz,
  Sundaram, Dodd, Fenimore, Gur, Petrick, Freymann, Kirby, Hughes, Casteele,
  Gupte, Sallam, Heath, Kuhn, Dharaiya, Burns, Fryd, Salganicoff, Anand,
  Shreter, Vastagh, Croft, and Clarke]{armato_iii_samuel_g._data_2015}
{Samuel G., Armato III}, M.~Geoffrey, B.~Luc, M.-G. Michael~F., M.~Charles~R.,
  R.~Anthony~P., Z.~Binsheng \emph{et~al.}, ``Data {{From LIDC}}-{{IDRI}},''
  2015.

\bibitem[Armato et~al.(2011)Armato, McLennan, Bidaut, McNitt-Gray, Meyer,
  Reeves, Zhao, Aberle, Henschke, Hoffman, et~al.]{armato_lung_2011}
S.~G. Armato, G.~McLennan, L.~Bidaut, M.~F. McNitt-Gray, C.~R. Meyer, A.~P.
  Reeves, B.~Zhao \emph{et~al.}, ``The lung image database consortium ({LIDC})
  and image database resource initiative ({IDRI}): a completed reference
  database of lung nodules on {CT} scans,'' \emph{Medical Physics}, vol.~38,
  no.~2, pp. 915--931, 2011.

\bibitem[Clark et~al.(2013)Clark, Vendt, Smith, Freymann, Kirby, Koppel, Moore,
  Phillips, Maffitt, Pringle, et~al.]{clark_cancer_2013}
K.~Clark, B.~Vendt, K.~Smith, J.~Freymann, J.~Kirby, P.~Koppel, S.~Moore
  \emph{et~al.}, ``The cancer imaging archive (tcia): maintaining and operating
  a public information repository,'' \emph{Journal of Digital Imaging},
  vol.~26, no.~6, pp. 1045--1057, 2013.

\bibitem[Setio et~al.(2016{\natexlab{a}})Setio, Ciompi, Litjens, Gerke, Jacobs,
  van Riel, Wille, Naqibullah, S{\'a}nchez, and van
  Ginneken]{setio_pulmonary_2016}
A.~A.~A. Setio, F.~Ciompi, G.~Litjens, P.~Gerke, C.~Jacobs, S.~J. van Riel,
  M.~M.~W. Wille \emph{et~al.}, ``Pulmonary nodule detection in {CT} images:
  false positive reduction using multi-view convolutional networks,''
  \emph{IEEE Transactions on Medical Imaging}, vol.~35, no.~5, pp. 1160--1169,
  2016.

\bibitem[Dou et~al.(2017)Dou, Chen, Yu, Qin, and Heng]{dou_multi-level_2016}
Q.~Dou, H.~Chen, L.~Yu, J.~Qin, and P.-A. Heng, ``Multilevel contextual 3-d
  cnns for false positive reduction in pulmonary nodule detection,'' \emph{IEEE
  Transactions on Biomedical Engineering}, vol.~64, no.~7, pp. 1558--1567,
  2017.

\bibitem[Setio et~al.(2016{\natexlab{b}})Setio, Traverso, de~Bel, Berens,
  Bogaard, Cerello, Chen, Dou, Fantacci, Geurts, et~al.]{setio_validation_2016}
A.~A.~A. Setio, A.~Traverso, T.~de~Bel, M.~S. Berens, C.~v.~d. Bogaard,
  P.~Cerello, H.~Chen \emph{et~al.}, ``Validation, comparison, and combination
  of algorithms for automatic detection of pulmonary nodules in computed
  tomography images: the luna16 challenge,'' \emph{arXiv preprint
  arXiv:1612.08012}, 2016.

\bibitem[Dundar et~al.(2008)Dundar, Fung, Krishnapuram, and
  Rao]{dundar_multiple-instance_2008}
M.~M. Dundar, G.~Fung, B.~Krishnapuram, and R.~B. Rao, ``Multiple-instance
  learning algorithms for computer-aided detection,'' \emph{IEEE Transactions
  on Biomedical Engineering}, vol.~55, no.~3, pp. 1015--1021, 2008.

\bibitem[Xu et~al.(2014)Xu, Mo, Feng, Zhong, Lai, Eric, and Chang]{xu2014deep}
Y.~Xu, T.~Mo, Q.~Feng, P.~Zhong, M.~Lai, I.~Eric, and C.~Chang, ``Deep learning
  of feature representation with multiple instance learning for medical image
  analysis,'' in \emph{IEEE International Conference on Acoustics, Speech and
  Signal Processing}.\hskip 1em plus 0.5em minus 0.4em\relax IEEE, 2014, pp.
  1626--1630.

\bibitem[Wang et~al.(2016)Wang, Yan, Tang, Bai, and Liu]{wang2016revisiting}
X.~Wang, Y.~Yan, P.~Tang, X.~Bai, and W.~Liu, ``Revisiting multiple instance
  neural networks,'' \emph{arXiv preprint arXiv:1610.02501}, 2016.

\bibitem[Wu et~al.(2015)Wu, Yu, Huang, and Yu]{wu2015deep}
J.~Wu, Y.~Yu, C.~Huang, and K.~Yu, ``Deep multiple instance learning for image
  classification and auto-annotation,'' in \emph{Proceedings of the IEEE
  Conference on Computer Vision and Pattern Recognition}, 2015, pp. 3460--3469.

\bibitem[Pinheiro and Collobert(2015)]{pinheiro2015image}
P.~O. Pinheiro and R.~Collobert, ``From image-level to pixel-level labeling
  with convolutional networks,'' in \emph{Proceedings of the IEEE Conference on
  Computer Vision and Pattern Recognition}, 2015, pp. 1713--1721.

\bibitem[Sun et~al.(2016)Sun, Han, Liu, and
  Khodayari-Rostamabad]{sun_multiple_2016}
M.~Sun, T.~X. Han, M.-C. Liu, and A.~Khodayari-Rostamabad, ``Multiple instance
  learning convolutional neural networks for object recognition,'' in
  \emph{International Conference on Pattern Recognition}.\hskip 1em plus 0.5em
  minus 0.4em\relax IEEE, 2016, pp. 3270--3275.

\bibitem[Zeng and Ji(2015)]{zeng2015deep}
T.~Zeng and S.~Ji, ``Deep convolutional neural networks for multi-instance
  multi-task learning,'' in \emph{Proceedings of the 2015 IEEE International
  Conference on Data Mining}.\hskip 1em plus 0.5em minus 0.4em\relax IEEE
  Computer Society, 2015, pp. 579--588.

\bibitem[Oni{\'s}ko et~al.(2001)Oni{\'s}ko, Druzdzel, and
  Wasyluk]{onisko2001learning}
A.~Oni{\'s}ko, M.~J. Druzdzel, and H.~Wasyluk, ``Learning bayesian network
  parameters from small data sets: Application of noisy-or gates,''
  \emph{International Journal of Approximate Reasoning}, vol.~27, no.~2, pp.
  165--182, 2001.

\bibitem[Anand and Downs(2008)]{anand2008probabilistic}
V.~Anand and S.~M. Downs, ``Probabilistic asthma case finding: a noisy or
  reformulation,'' in \emph{AMIA Annual Symposium Proceedings}, vol.
  2008.\hskip 1em plus 0.5em minus 0.4em\relax American Medical Informatics
  Association, 2008, p.~6.

\bibitem[Heckerman(1990)]{heckerman1990tractable}
D.~Heckerman, ``A tractable inference algorithm for diagnosing multiple
  diseases,'' in \emph{Proceedings of the Fifth Annual Conference on
  Uncertainty in Artificial Intelligence}.\hskip 1em plus 0.5em minus
  0.4em\relax North-Holland Publishing Co., 1990, pp. 163--172.

\bibitem[Halpern and Sontag(2013)]{halpern2013unsupervised}
Y.~Halpern and D.~Sontag, ``Unsupervised learning of noisy-or bayesian
  networks,'' \emph{arXiv preprint arXiv:1309.6834}, 2013.

\bibitem[Zhang et~al.(2006)Zhang, Platt, and Viola]{zhang2006multiple}
C.~Zhang, J.~C. Platt, and P.~A. Viola, ``Multiple instance boosting for object
  detection,'' in \emph{Advances in Neural Information Processing Systems},
  2006, pp. 1417--1424.

\bibitem[MacMahon et~al.(2017)MacMahon, Naidich, Goo, Lee, Leung, Mayo, Mehta,
  Ohno, Powell, Prokop, et~al.]{macmahon2017guidelines}
H.~MacMahon, D.~P. Naidich, J.~M. Goo, K.~S. Lee, A.~N. Leung, J.~R. Mayo,
  A.~C. Mehta \emph{et~al.}, ``Guidelines for management of incidental
  pulmonary nodules detected on ct images: from the fleischner society 2017,''
  \emph{Radiology}, p. 161659, 2017.

\bibitem[Ronneberger et~al.(2015)Ronneberger, Fischer, and
  Brox]{ronneberger_u-net_2015}
O.~Ronneberger, P.~Fischer, and T.~Brox, ``U-net: Convolutional networks for
  biomedical image segmentation,'' in \emph{International Conference on Medical
  Image Computing and Computer-Assisted Intervention}.\hskip 1em plus 0.5em
  minus 0.4em\relax Springer, 2015, pp. 234--241.

\bibitem[He et~al.(2016)He, Zhang, Ren, and Sun]{he_deep_2015}
K.~He, X.~Zhang, S.~Ren, and J.~Sun, ``Deep residual learning for image
  recognition,'' in \emph{Proceedings of the IEEE Conference on Computer Vision
  and Pattern Recognition}, 2016, pp. 770--778.

\bibitem[Girshick et~al.(2014)Girshick, Donahue, Darrell, and
  Malik]{girshick_rich_2014}
R.~Girshick, J.~Donahue, T.~Darrell, and J.~Malik, ``Rich feature hierarchies
  for accurate object detection and semantic segmentation,'' in
  \emph{Proceedings of the IEEE Conference on Computer Vision and Pattern
  Recognition}, 2014, pp. 580--587.

\bibitem[Ioffe and Szegedy(2015)]{ioffe2015batch}
S.~Ioffe and C.~Szegedy, ``Batch normalization: Accelerating deep network
  training by reducing internal covariate shift,'' in \emph{International
  Conference on Machine Learning}, 2015, pp. 448--456.

\bibitem[Ha and Mazzone()]{ha2014pulmonary}
D.~M. Ha and P.~J. Mazzone, ``Pulmonary nodules,'' \emph{Age}, vol.~30, pp.
  0--05.

\bibitem[He et~al.(2017)He, Gkioxari, Doll{\'a}r, and Girshick]{he_mask_2017}
K.~He, G.~Gkioxari, P.~Doll{\'a}r, and R.~Girshick, ``Mask {R-CNN},'' in
  \emph{International Conference on Computer Vision}, 2017.

\end{thebibliography}

\end{document}